\documentclass[10pt,twocolumn,letterpaper]{article}

\usepackage{iccv}
\usepackage{times}
\usepackage{epsfig}
\usepackage{graphicx}
\usepackage{amsmath}
\usepackage{amssymb}
\usepackage{comment}
\usepackage[justification=justified,singlelinecheck=false]{caption}
\usepackage{booktabs,siunitx}
\usepackage{multirow}
\usepackage{xcolor}
\usepackage{url}

\usepackage{textcomp}
\usepackage{wrapfig}
\usepackage{microtype}
\newtheorem{theorem}{Theorem}
\newcommand\blfootnote[1]{%
  \begingroup
  \renewcommand\thefootnote{}\footnote{#1}%
  \addtocounter{footnote}{-1}%
  \endgroup
}

\iccvfinalcopy 

\def\iccvPaperID{1442} 

\ificcvfinal\fi
\newcommand{\normal}{\mathbf{n}}
\newcommand{\tangent}{\mathbf{t}}
\newcommand{\bitangent}{\mathbf{b}}
\newcommand{\upvector}{\mathbf{u}}

\newcommand{\camF}{\mathbf{F}^{c}}
\newcommand{\upF}{\mathbf{F}^{g}}

\newcommand{\upFz}{\mathbf{f}_z^g}

\newcommand{\Lgrad}{\mathcal{L}_{\nabla}}
\newcommand{\Lpose}{\mathcal{L}_{o}}
\newcommand{\Lgeo}{\mathcal{L}_{F}}

\newenvironment{packed_item}{
\begin{itemize}
  \setlength{\itemsep}{1pt}
  \setlength{\parskip}{2pt}
  \setlength{\parsep}{0pt}
}{\end{itemize}}

\begin{document}

\title{UprightNet: Geometry-Aware Camera Orientation Estimation \\
from Single Images}

\author{Wenqi Xian${}^{*,1}$\
Zhengqi Li${}^{*,1}$\ 
Matthew Fisher${}^{2}$\ 
Jonathan Eisenmann${}^{2}$\ 
Eli Shechtman${}^{2}$\ 
Noah Snavely${}^{1}$
\\[2mm]
${}^{1}$ Cornell Tech, Cornell University\qquad 
${}^{2}$ Adobe Research
}

\maketitle

\begin{abstract}
We introduce \emph{UprightNet}, a learning-based approach for estimating 2DoF camera orientation from a single RGB image of an indoor scene. Unlike recent methods that leverage deep learning to perform black-box regression from image to orientation parameters, we propose an end-to-end framework that incorporates explicit geometric reasoning. In particular, we design a network that predicts two representations of scene geometry, in both the local camera and global reference coordinate systems, and solves for the camera orientation as the rotation that best aligns these two predictions via a differentiable least squares module. This network can be trained end-to-end, and can be supervised with both ground truth camera poses and intermediate representations of surface geometry. We evaluate UprightNet on the single-image camera orientation task on synthetic and real datasets, and show significant improvements over prior state-of-the-art approaches.
\end{abstract}

\blfootnote{$^{*}$ indicates equal contribution}
\section{Introduction}

\begin{figure}[t]
\begin{center}
  \includegraphics[width=\columnwidth]{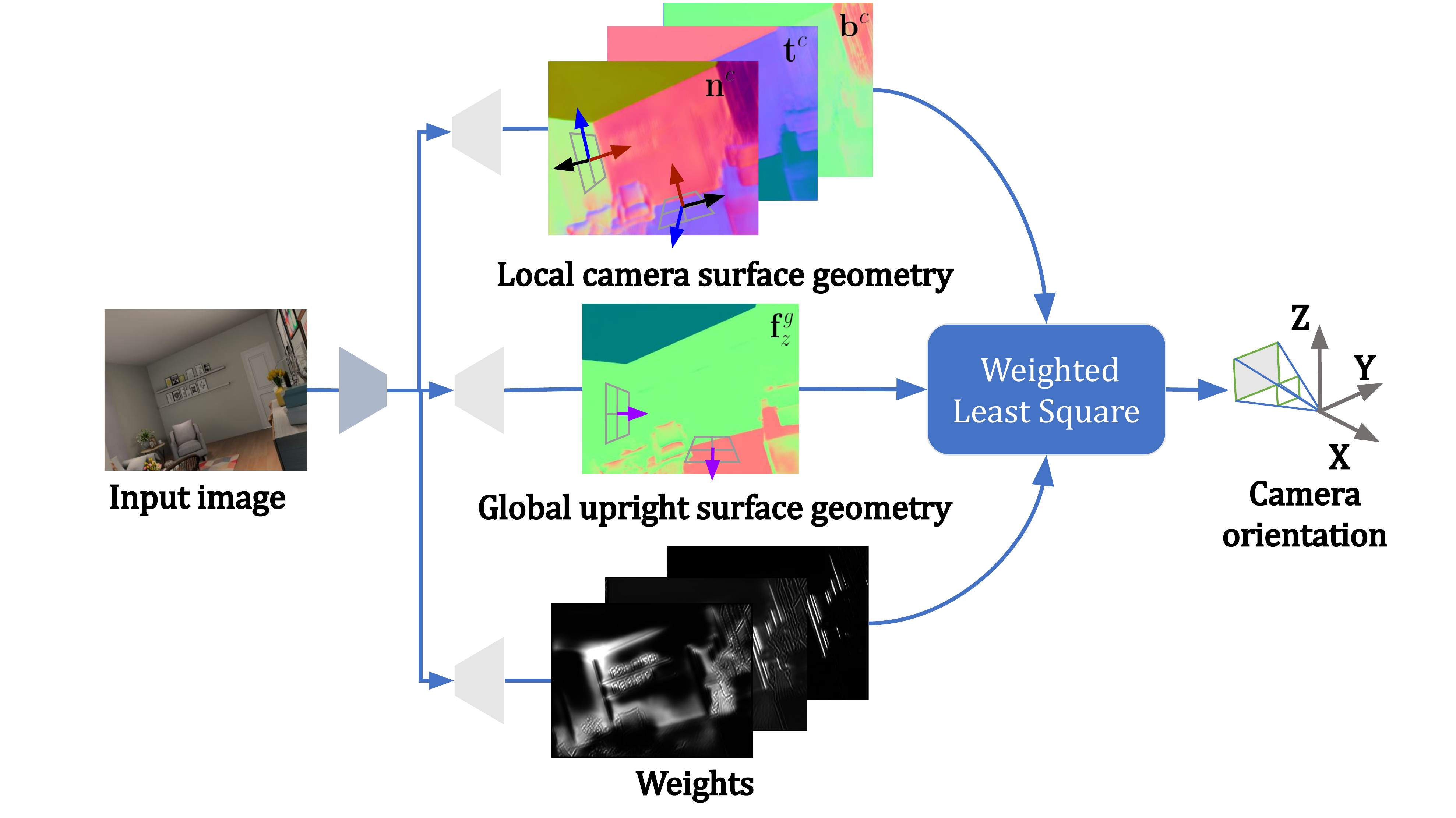}
\end{center}
  \caption{\textbf{UprightNet overview.} UprightNet takes a single RGB image and predicts surface geometry in both \textit{local camera} and \textit{global upright} coordinate systems. The camera orientation is then computed as the \emph{alignment} between these two predictions, solved for by a differentiable least squares module, and weighted using predicted weight maps.} 
\label{fig:pipeline}
\end{figure}

We consider the problem of estimating camera orientation from a single RGB photograph. This is an important problem with applications in robotics, image editing, and augmented reality.
Classical approaches often rely on 2D projective geometric cues such as vanishing points~\cite{lee2014automatic}.
However, more recent methods have sought to leverage the power of deep learning to directly regress from an image to extrinsic calibration parameters, by training on images with known ground truth orientation information~\cite{workman2016hlw,hold2018perceptual}.
But these methods typically do not explicitly leverage the knowledge of projective geometry, treating the problem as a black-box regression or classification.

In this work, we introduce UprightNet, a novel deep network model for extrinsic camera calibration that incorporates explicit geometric principles.
We hypothesize that injecting geometry will help achieve better performance and better generalization to a broader class of images, because geometry affords generally applicable principles, and because geometric representations provide a structured intermediary in the otherwise highly non-linear relationship between raw pixels and camera orientation.


In particular, we define and use \emph{surface frames} as an intermediate geometric representation. The orthogonal basis of the surface frames include the surface normal and two vectors that span the tangent plane of the surface. Surface frames allow us to capture useful geometric features---for instance, predicted surface normals on the ground will point directly in the up direction, and horizontal lines in the scene will point perpendicular to the up direction. However, it is not enough to know normals and other salient vectors in camera coordinates, we also need to know \textit{which normals are on the ground}, etc. Therefore, our insight is to predict surface geometry not only in local camera coordinates, but also in \emph{global upright} coordinates, as shown in Figure~\ref{fig:pipeline}. 

Such a global prediction is consistent across different camera views and is highly related to the semantic task of predicting which pixels are horizontal surfaces (floors and ceilings), and which are vertical (walls). The camera orientation can then be estimated as the rotation that best aligns these two representations of surface geometry. This overall approach is illustrated in Figure~\ref{fig:pipeline}.
This alignment problem can be solved as a constrained least squares problem. We show in this paper that such an approach is end-to-end differentiable, allowing us to train the entire network using both supervisions on the intermediate representation of surface geometry, as well as on the final estimated orientation. 






We evaluate UprightNet on both synthetic and real-world RGBD datasets, and compare it against classical geometry-based and learning-based methods. Our method shows significant improvements over prior methods, suggesting that the geometry guidance we provide to the network is important for learning better camera orientation.



\section{Related Work}


Single image camera calibration is a longstanding problem in computer vision. Classical geometry-based methods highly rely on low-level image cues. When only a single image is available, parallel and mutually orthogonal line segments detected in the images can be used to estimate vanishing points and the horizon lines~\cite{joo2016globally, miyagawa2010simple, deutscher2002automatic, lee2014automatic, mirzaei2011optimal, bazin2012globally, xu2013minimum, bazin20123, lezama2014finding, wildenauer2012robust, xu2017pose, antunes2013global}. Other techniques based on the shape of objects such as coplanar circles~\cite{chen2004camera} and repeated patterns~\cite{schaffalitzky2000planar} have also been proposed. When an RGB-D camera is available, one can solve for the upright orientation by assuming an ideal Manhattan world~\cite{ghanem2015robust, straub2015real, straub2018manhattan, kim2018indoor}. When the scene in question has been mapped in 3D, one can also solve for the camera pose by re-localizing the cameras with respect to the 3D maps~\cite{li2010location,brachmann2017dsac, sattler2011fast,brachmann2018learning,kendall2015posenet}. 

On the other hand, using machine learning methods to estimate camera orientation from a single image has gained attention. Earlier work proposes to detect and segment skylines of the images in order to estimate horizon lines~\cite{fefilatyev2006horizon, ahmad2013machine}. More recently, CNN-based techniques have been developed for horizon estimation from a single image~\cite{zhai2016detecting,hold2018perceptual,workman2016hlw}. Most of these methods formulate the problem as either regression or classification and impose a strong prior on the location of features correlated with the visible horizon and of corresponding camera parameters.




Single-image surface normal prediction powered by deep networks~\cite{wang2015designing, eigen2015predicting} can provide a supervision signal for many 3D vision tasks such as planar reconstruction~\cite{liu2018planenet}, depth completion~\cite{zhang2018deep} and 2D-3D alignment \cite{bansal2016marr}. Recently,
surface normal was used for single-image camera estimation by directly estimating a ground plane from the depth and normal estimates of segmented ground regions~\cite{man2018groundnet}. Unfortunately, such method assumes the ground plane is always visible in the images, and only applies to vehicle-control use cases. In addition, there are recent work making use of local surface frame representation for a variety of 3D scene understanding tasks~\cite{huang2019texturenet, huang2019framenet}. However, our method extends beyond these ideas by estimating both local and \textit{global aligned} surface geometry from single images and use such correspondences to estimate camera orientation. Suwajanakorn~\etal~\cite{suwajanakorn2018discovery} shows that the supervision on relative pose could automatically discover consistent keypoints of 3D objects, and our method is inspired by this work on end-to-end learning of intermediate representation via pose supervision. 





\section{Approach}

Man-made indoor scenes typically consist of prominently structured surfaces such as walls, floors, and ceilings, as well as lines and junctions arising from intersections between these structures. Prominent lines also arise from other scene features, such as oriented textures. In indoor images, such geometric cues provides rich information about camera orientation. We propose to exploit such geometric features by explicitly predicting surface geometry as an intermediate step in estimating camera orientation. 

To understand the benefits of such an approach, imagine that we take an image and predict per-pixel surface normals, in camera coordinates. How do these relate to camera orientation? Surface normals on the ground and other horizontal surfaces point in the same direction as the camera’s up vector---exactly the vector we wish to estimate. Similarly, surface normals on walls and other vertical surfaces are perpendicular to the up vector. Hence, finding the camera orientation can be posed as finding the vector that is most parallel to ground normals, and at the same time most perpendicular to wall normals. 

However, such an approach assumes that we know which pixels are ground and which are walls. Thus, we propose to also predict normals in \emph{global scene coordinates}. This approach is in contrast to most work that predicts surface normals, which usually predict them only in the camera reference frame (e.g., if the camera is rolled 45 degrees about its axis, the predicted normals should be rotated accordingly). Given surface geometry predicted in \emph{both} camera and scene coordinates, the camera orientation can be found as the rotation that best \emph{aligns} these two frames.

This approach is suitable for learning. If the alignment procedure is differentiable, then we can train a method end-to-end to predict orientation by comparing to ground-truth orientations. A key advantage is that we can also apply supervision to the intermediate geometric predictions if we have ground truth.

What kind of surface geometry should we estimate? Normals are useful as described above, but do not capture in-plane features such as junctions and other lines. Hence, we propose to estimate a full orthonormal coordinate frame at each pixel, comprised of a normal and two tangent vectors. We predict these frames as a dual surface geometry representation in two coordinate systems:
\begin{packed_item}
  \item $\camF$: the surface geometry in \textit{local camera} coordinates.
  \item $\upF$: the surface geometry in \textit{global upright} coordinates. 
\end{packed_item}

\begin{figure}[t]
    \begin{tabular}{@{\hspace{-0.1em}}c@{\hspace{-0.1em}}c@{\hspace{-0.1em}}c@{\hspace{-0.1em}}c@{\hspace{-0.1em}}c@{\hspace{-0.1em}}}
        \includegraphics[width=0.19\columnwidth]{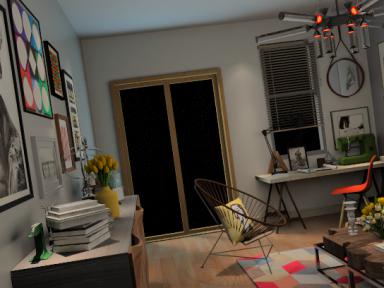} \vspace{-0.05em} & 
        \includegraphics[width=0.19\columnwidth]{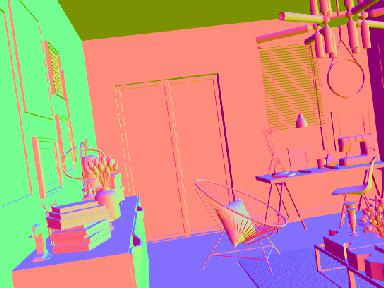}  \vspace{-0.05em} &
        \includegraphics[width=0.19\columnwidth]{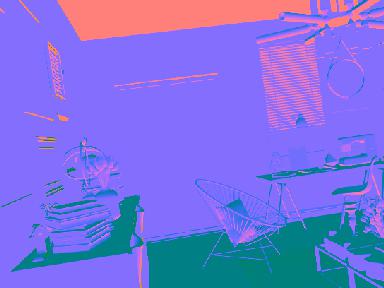}  \vspace{-0.05em} &
        \includegraphics[width=0.19\columnwidth]{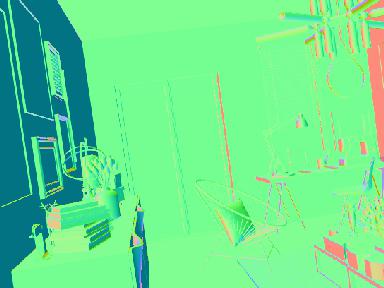}  \vspace{-0.05em} &
        \includegraphics[width=0.19\columnwidth]{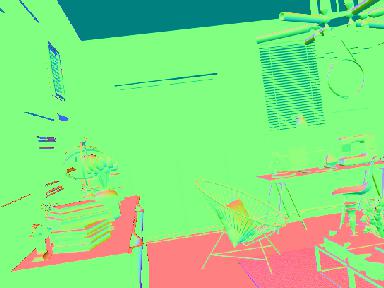} \vspace{-0.05em} \\ 
        \includegraphics[width=0.19\columnwidth]{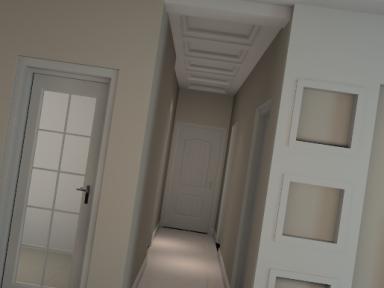} \vspace{-0.05em} & 
        \includegraphics[width=0.19\columnwidth]{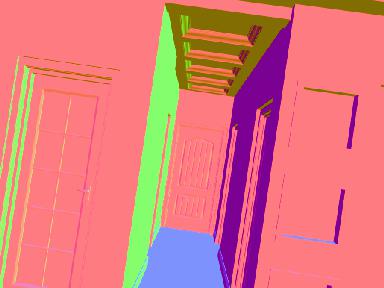}  \vspace{-0.05em} &
        \includegraphics[width=0.19\columnwidth]{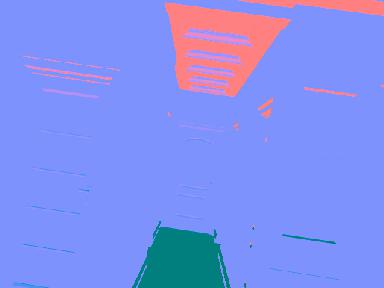}  \vspace{-0.05em} &
        \includegraphics[width=0.19\columnwidth]{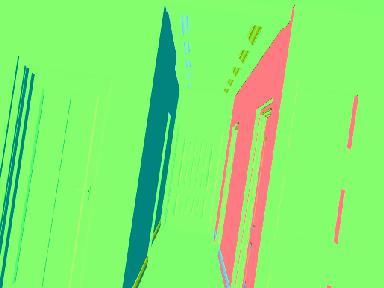}  \vspace{-0.05em} &
        \includegraphics[width=0.19\columnwidth]{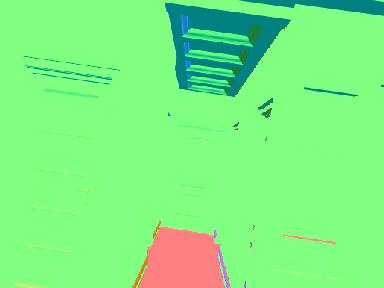} \vspace{-0.05em} \\ 
        \includegraphics[width=0.19\columnwidth]{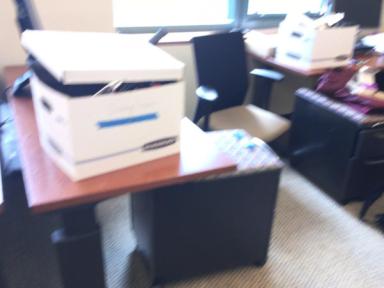} \vspace{-0.05em} & 
        \includegraphics[width=0.19\columnwidth]{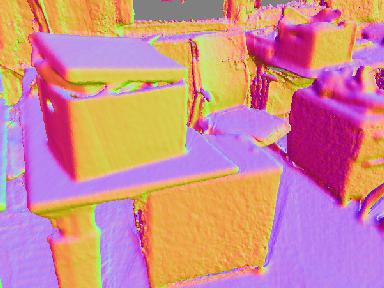}  \vspace{-0.05em} &
        \includegraphics[width=0.19\columnwidth]{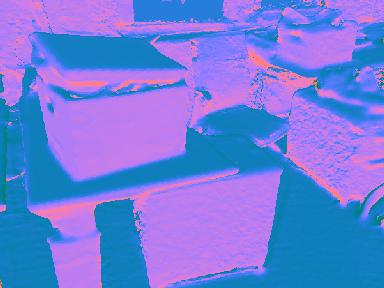}  \vspace{-0.05em} &
        \includegraphics[width=0.19\columnwidth]{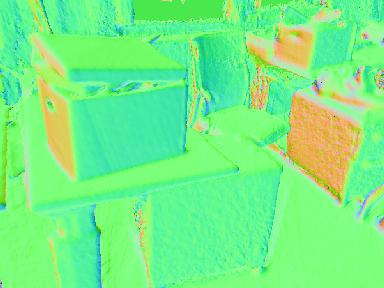}  \vspace{-0.05em} &
        \includegraphics[width=0.19\columnwidth]{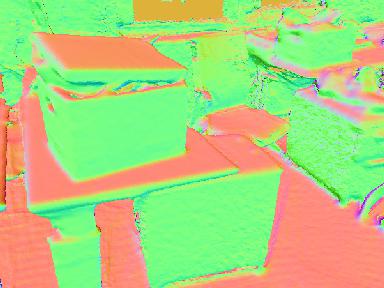} \vspace{-0.05em} \\ 
        \includegraphics[width=0.19\columnwidth]{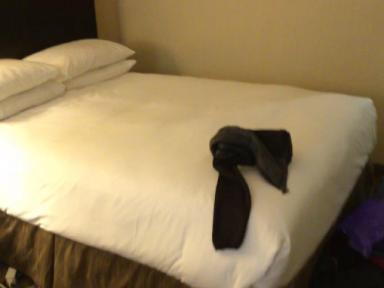} \vspace{-0.05em} & 
        \includegraphics[width=0.19\columnwidth]{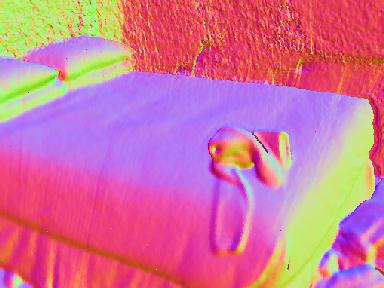}  \vspace{-0.05em} &
        \includegraphics[width=0.19\columnwidth]{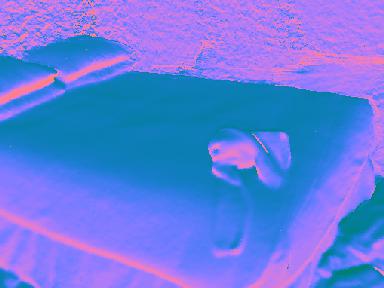}  \vspace{-0.05em} &
        \includegraphics[width=0.19\columnwidth]{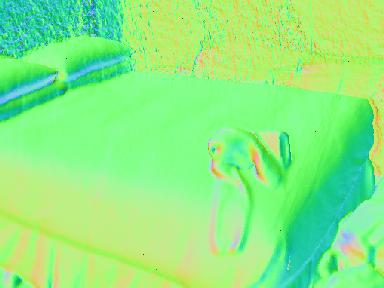}  \vspace{-0.05em} &
        \includegraphics[width=0.19\columnwidth]{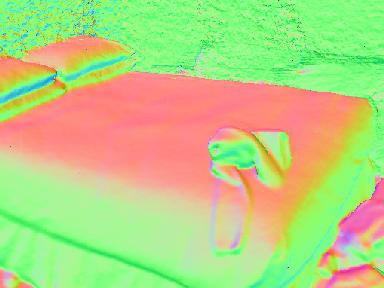} \vspace{-0.05em} \\ 
        {\small (a) Image} & {\small (b) $\normal^{c}$} & {\small (c) $\tangent^{c}$} & {\small (d) $\bitangent^{c}$} & {\small (e) $\upFz$}
    \end{tabular} 
  \caption{\textbf{Visualization of surface geometry.} From left to right: (a) image, (b-d) local camera surface frames $\camF$, (e) the third row of global upright surface frames $\upF$.}\label{fig:viz_sur_geo}
\end{figure}

\noindent \textbf{Surface frames.}
To represent surface geometry, we define a surface frame $\mathbf{F}(i)$ at every pixel location $i$ as a $3\times3$ matrix formed by three mutually orthogonal unit vectors $\mathbf{F}(i) = \left[ \normal(i)\  \tangent(i)\ \bitangent(i) \right]$ in either local camera coordinates or global upright coordinates:
\begin{packed_item}
  \item $\normal^{c}$, $\normal^{g}$: surface normal in camera and upright coordinates, respectively.
  \item $\tangent^{c}$, $\bitangent^{c}$, $\tangent^{g}$, $\bitangent^{g}$: mutually orthogonal unit vectors that span the tangent plane of the corresponding surface normal, in camera and upright coordinates respectively. ($\tangent$ stands for tangent, and $\bitangent$ for bitangent.)
\end{packed_item}
As usual, we define the camera coordinate system as a view-dependent local coordinate system, and the global upright coordinate system as the one whose camera up vector aligns with global scene up vector. 

Which tangent vectors should we choose for $\tangent$ and $\bitangent$? For curved surfaces, these tangent vectors are often defined in terms of local curvature. However, for man-made indoor scenes, many surfaces are planar and hence lack curvature. Instead, we define these vectors to align with the upright orientation of the scene. In particular, we define the tangent vector $\tangent$ as a unit vector derived from the cross product between the surface normal $\normal$ and the camera $y$-axis (pointing rightward in our case). The bitangent vector $\bitangent$ is then $\bitangent = \normal \times \tangent$.
This definition of tangent vectors has a degeneracy when the surface normal is parallel to the camera $y$-axis, in which case we instead compute $\tangent$ to align with the up vector. However, an advantage of this choice of tangents is that this degeneracy is rare in practice. We find this choice leads to the best performance in our experiments. However, other surface frame representations could also be used~\cite{huang2019texturenet,huang2019framenet}.


\medskip
\noindent \textbf{Camera orientation.}
Let $\mathbf{R}$ be the $3\times 3$ rotation matrix transforming local camera coordinates to global upright coordinates. $\mathbf{R}$ maps an upright surface frame $\upF(i)$ to its corresponding camera surface frame  $\camF(i)$ as follows:
\begin{equation}
  \upF(i) = \mathbf{R} \camF(i)
\end{equation}
Note that there is no natural reference for determining the camera heading (yaw angle) from a single image, and moreover we are most interested in determining camera roll and pitch because they are useful for graphics applications. Therefore, our problem is equivalent to finding the scene up vector in the camera coordinate system, which we denote $\upvector$, and which happens to be the same as the third row of $\mathbf{R}$. The scene up vector $\upvector$ encodes both roll and pitch, but not yaw. It also relates the two surface frames as follows:
\begin{equation}
    \upFz(i) = \upvector^T \camF(i) \label{eq:ls_1}
\end{equation}
where we define $\upFz(i) = [ \normal^g_z(i) \ \tangent^g_z(i) \ \bitangent^g_z(i) ] \in \mathbb{R}^{3}$ as the third row of $\upF(i)$ and it has unit length by definition. 

The last column of Figure~\ref{fig:viz_sur_geo} shows the vectors $\upFz(i)$. Note that $\upFz(i)$ is consistent in the same supporting surfaces across images, and hence we refer to it as a \emph{scene layout vector}. For example, $n_z^g$ for ground, wall and ceiling pixels is always fixed, to $1$, $0$ and $ \text{-}1$, respectively, across \emph{all} images, while they can differ in camera coordinates for different images according to camera orientation. Therefore, a beneficial property of the global upright frame representation is that it is similar in spirit to performing a semantic segmentation of the ground, ceiling, and other supporting structures. 





To estimate 3DoF camera orientation, we could predict both camera and upright surface frames for an image, then estimate a rotation matrix that best aligns these frames. However, 
since we only estimate 2DoF camera orientation, it is sufficient to predict $\camF$ and $\upFz$.




Figure \ref{fig:pipeline} shows an overview of our approach. Given a single RGB image, our network predicts per-pixel local camera surface frames $\camF$ and scene layout vectors $\upFz$.
Using corresponding local/upright frames, we can formulate computing the best up vector as a 
constrained least squares problem. We show how this problem can be solved in a differentiable manner (Sec.~\ref{pose}), allowing us to train a network end-to-end by supervising it with ground truth camera orientations.

\medskip
\noindent \textbf{Predicting weights.}
A key challenge in our problem formulation is the varying uncertainty of surface geometry predictions in different image regions. We solve orientation estimation via rigid alignment as a least squares problem, which is sensitive to outliers in the predicted surface frames.

To address this problem, at each pixel $i$, we propose to additionally predict separate weights $w_\mathbf{\normal}(i)$,  $w_\mathbf{\tangent}(i)$,  $w_\mathbf{\bitangent}(i)$ for each of the $\normal$, $\tangent$ and $\bitangent$ maps, and integrate these weights into the least squares solver. We have no ground truth weights available for supervision, but because we can train our system end-to-end, the network can learn by itself to focus on only the most reliable predicted regions. Hence during training, our model jointly optimizes for surface frames, weights maps, and camera orientation.

\subsection{Up vector from surface frame correspondences} \label{pose}

\noindent \textbf{Differentiable constrained least squares.}
Given local surface frames $\camF$ and the corresponding $\upFz$, our goal is to find the up vector $\upvector$ that best aligns them. Given Eq.~\ref{eq:ls_1}, we can write the following constrained minimization problem:
\begin{align}
    \min_{\mathbf{\upvector}} & \sum_{i=1}^N  \left\Vert \upvector^T  \camF(i) - \upFz(i) \right\Vert_2^2 \nonumber \\ 
   & \text{subject to } \left\Vert \upvector \right\Vert_2 = 1 \label{eq:cls_1}
\end{align}
where $N$ is the number of pixels. Eq.~\ref{eq:cls_1} can be rewritten in matrix form as:
\begin{align}
    \min_{\upvector} \left\Vert \mathbf{A}\upvector - \mathbf{b} \right\Vert_2^2, \ \ 
    \text{subject to } \left\Vert \upvector \right\Vert_2 = 1  \label{eq:cls_2}
\end{align}
where the matrix $\mathbf{A} \in \mathbb{R}^{3N\times 3}$ can be formed by vertically stacking matrices $\camF(i)$ for each pixel $i$, and similarly vector $\mathbf{b} \in \mathbb{R}^{3N}$ can be formed by stacking the vectors $\upFz(i)$.

If there were no unit-norm constraint, this problem would be a standard least squares problem. Similarly, if $\mathbf{b} = \mathbf{0}$, the problem becomes a homogeneous least squares problem that can be solved in closed form using SVD~\cite{hartley2003multiple}. In our case, $\mathbf{b}$ is not necessarily a zero vector, preventing us using such standard approaches. However, we show that Eq.~\ref{eq:cls_2} can be solved analytically, allowing us to use it to compute a loss in an end-to-end training pipeline.

In particular, we can write the Lagrangian of Eq.~\ref{eq:cls_2} as 
\begin{equation}
    L = (\mathbf{A}\upvector - \mathbf{b})^T (\mathbf{A}\upvector - \mathbf{b}) - \lambda (\upvector^T \upvector - 1) \label{eq:lagrangian_1}
\end{equation}
where $\lambda$ is a Lagrange multiplier. The Karush–Kuhn–Tucker condition of Eq.~\ref{eq:lagrangian_1} leads to the following equations:
\begin{equation}
    (\mathbf{A}^T \mathbf{A} - \lambda \mathbf{I}) \upvector = \mathbf{A}^T \mathbf{b}, \quad \upvector^T\upvector=1 \label{eq:cls_4}
\end{equation}
To solve for $\lambda$ and $\upvector$ from Eq.~\ref{eq:cls_4} analytically, we use the techniques proposed in \cite{gander1989constrained}. Specifically, we have following theorem~\cite{gander1989constrained}: 
\begin{theorem}
Eq.~\ref{eq:cls_4} can be reduced to a quadratic eigenvalue problem (QEP):
\begin{equation}
    \mathbf{I} \lambda^2 - 2 \mathbf{H} \lambda + \mathbf{H}^2 - \mathbf{g}\mathbf{g}^T = 0 \label{eq:qep_1}
\end{equation}
where $\mathbf{H}=\mathbf{A}^T \mathbf{A}$, $\mathbf{g}=\mathbf{A}^T \mathbf{b}$, and Eq.~\ref{eq:qep_1} has a solution for $\lambda$. Further, the solution $\lambda$ and $\upvector = \left( \mathbf{H} - \lambda \mathbf{I}\right)^{-1} \mathbf{g}$ satisfies $(H - \lambda \mathbf{I}) \upvector = \mathbf{g} $ and $\upvector^T \upvector = 1$. 

\end{theorem}

We refer readers to the supplementary material and to \cite{gander1989constrained} for the proof. 
Fortunately, to solve this QEP, we can reduce it to an ordinary eigenvalue problem~\cite{gander1989constrained}:
\begin{equation}
  \begin{bmatrix}
    \mathbf{H} & -\mathbf{I} \\
    \mathbf{- \mathbf{g} \mathbf{g}^T} & \mathbf{H} 
  \end{bmatrix} 
  \begin{bmatrix}
  \gamma \\ \mu 
  \end{bmatrix}  = 
  \lambda   \
  \begin{bmatrix}
  \gamma \\ \mu 
  \end{bmatrix} \label{eq:eigen_1}
\end{equation}
where $\gamma = (\mathbf{H} - \lambda \mathbf{I})^{-2} \mathbf{g}$ and $\mu = (H - \lambda \mathbf{I}) \gamma$. Since the block matrix on the left hand side of Eq.~\ref{eq:eigen_1} is not necessarily symmetric, the optimal $\lambda$ corresponds to its minimum real eigenvalue.
The derivative of this eigenvalue can be found in closed form~\cite{van2005perturbation}, and so the solver is fully differentiable.


\medskip
\noindent \textbf{Weighted least squares.}
To improve the robustness of the least squares solver, we weight each correspondence in Eq.~\ref{eq:cls_1}:
\begin{align}
    \min_{\mathbf{\upvector}} & \sum_{i=1}^N  \left\Vert \mathbf{W}(i) \left( \upvector^T  \camF(i) - \upFz(i) \right)^T   \right\Vert_2^2 \nonumber \\
    & \text{subject to } \left\Vert \upvector \right\Vert_2 = 1 \label{eq:cls_w1}
\end{align}
and corresponding Lagrangian can be similarly modified as 
\begin{equation}
    L' = (\mathbf{A}\upvector - \mathbf{b})^T \mathbf{W}^T \mathbf{W} (\mathbf{A}\upvector - \mathbf{b}) - \lambda (\upvector^T \upvector - 1) \label{eq:cls_w2}
\end{equation}
where $\mathbf{W} \in \mathbb{R}^{3N \times 3N}$ is a diagonal matrix, and each $3 \times 3$ block, denoted as $\mathbf{W}(i)$, is $\text{diag}([w_{\normal}(i)\  w_{\tangent}(i)\ w_{\bitangent}(i)])$. Hence, we can use the technique described above to solve for $\lambda$ and $\upvector$. In our experiments, we show that the predicted weights not only help to reduce the overall estimation error in the presence of noisy predictions, but also focus on supporting structures, as shown in Figure~\ref{fig:interiornet_pred} and Figure~\ref{fig:alignment}.

\subsection{Loss functions}

UprightNet jointly optimizes for surface frames, weights, and camera orientation in an end-to-end fashion. Our overall loss function is the weighted sum of terms: 
\begin{equation}
    \mathcal{L}_{\text{total}} = \Lpose + \alpha_{F} \Lgeo  + 
    \alpha_{\nabla} \Lgrad
\end{equation}
In contrast to prior approaches that directly perform regression or classification on the ground-truth camera orientation, our method explicitly makes use of geometric reasoning over the entire scene, and we can train a network end-to-end with two primary objectives: 
\begin{packed_item}
  \item[$\bullet$] A \textbf{camera orientation loss} that measures the error between recovered up vector and the ground-truth.  
  \item[$\bullet$] A \textbf{surface geometry loss} that measures errors between predicted surface frames and ground-truth surface frames in both local camera and global upright coordinate systems.
\end{packed_item}   

\noindent \textbf{Camera orientation loss $\Lpose$.} 
The camera orientation loss is applied to the up vector estimated by the surface frame correspondences and weights using our proposed constrained weighted least squares solver. Specifically, the loss is defined as the angular distance between the estimated up vector $\hat{\upvector}$ and the ground-truth one $\upvector$: 
\begin{equation}
  \label{eq:4}
  \Lpose =
    \text{arccos}\left(\hat{\upvector} \cdot \upvector \right) 
\end{equation}
Note that both $\hat{\upvector}$ and $\upvector$ are unit vectors. We can backpropagate through our differentiable constrained weighted least squares solver to minimize this loss directly.

A numerical difficulty is that the gradient of $\operatorname{arccos}(x)$ reaches infinity when $x=1$. To avoid exploding gradients, our loss automatically switches to $1 - \hat{\upvector}^T \upvector$ when $\hat{\upvector} \cdot \upvector$ is greater than $1 - \epsilon$. In our experiments, we set $\epsilon={10}^{-6}$ and find that this strategy leads to faster training convergence and better performance compared to alternatives.

\medskip
\noindent \textbf{Surface frames loss $\Lgeo$.} We also introduce a supervised loss $\Lgeo$ over predicted surface frames in both coordinate systems to encourage the network to learn a consistent surface geometry representation. In particular, we compute the cosine similarity between each column of $\hat{\mathbf{F}}^c$ and the corresponding column of the ground-truth $\camF$. We also compute the cosine similarity between $\hat{\mathbf{f}}_z^g$ and the ground-truth $\upFz$,
yielding the following loss:
\begin{equation}
    \small
    \Lgeo = 2 - \frac{1}{3N}\sum_{i=1}^N \sum_{\mathbf{f} \in \{ \normal, \tangent, \bitangent \}}  {\hat{\mathbf{f}}^c(i) \cdot \mathbf{f}^c(i) - \frac{1}{N}\sum_{i=1}^N \hat{\mathbf{f}}_z^g(i) \cdot \upFz(i)}
\end{equation}


\medskip
\noindent \textbf{Gradient consistency loss $\Lgrad$.} Finally, to encourage piecewise constant predictions on flat surfaces and sharp discontinuities, we include a gradient consistency loss across multiple scales, similar to prior work~\cite{li2019learning,li2018cgintrinsics,li2018megadepth}. The gradient consistency loss $\Lgrad$ measures the $\ell_1$ error between the gradients of the prediction and the corresponding ground truth:
\begin{align}
    \small
     \Lgrad = \sum_{s=1}^S \frac{1}{3N_s}\sum_{i=1}^{N_s} \sum_{\mathbf{f} \in \{ \normal, \tangent, \bitangent \}} 
     || \nabla \hat{\mathbf{f}}^c(i)- \nabla\mathbf{f}^c(i) ||_1 \nonumber \\ 
     + \sum_{s=1}^S \frac{1}{N_s}\sum_{i=1}^{N_s} || \nabla \hat{\mathbf{f}}_z^g(i)- \nabla \upFz(i) ||_1 
\end{align}
where $S$ is the number of scales, and $N_s$ is the number of pixels in each scale. In our experiments, we set $S=4$ and use nearest neighborhood downsampling to create image pyramids for both the prediction and ground-truth. 



\subsection{Network architecture}

We adopt a U-Net-style network architecture~\cite{godard2017unsupervised, ronneberger2015u} for UprightNet. Our network consists of one encoder and three separate decoders for $\hat{\mathbf{F}}^c$ (9 channels), $\hat{\mathbf{f}}_z^g$ (3 channels) and weight maps (3 channels), respectively. We adopt an ImageNet~\cite{russakovsky2015imagenet} pretrained ResNet-50~\cite{he2016deep} as the backbone encoder. Each decoder layer is composed of a $3 \times 3$ convolutional layer followed by bilinear upsampling, and skip connections are also applied. We normalize each column of $\hat{\mathbf{F}}^c$ and $\hat{\mathbf{f}}_z^g$ to unit length. For the weight maps, we add a sigmoid function at the end of the weight stream and normalize predicted weight maps by dividing them by their mean.

\section{Experiments}

To validate the effectiveness of UprightNet, we train and test on synthetic images from the InteriorNet dataset~\cite{li2018interiornet} and real data from ScanNet~\cite{dai2017scannet}, and compare with several prior single-image orientation estimation methods. Furthermore, to test generalization ability, we directly apply all methods trained on ScanNet to images drawn from the SUN360~\cite{xiao2012recognizing} dataset without fine-tuning. For all datasets, we show both qualitative and quantitative results, as well as comparisons to other baselines.





\subsection{Datasets}

\noindent
\textbf{InteriorNet~\cite{li2018interiornet}} is a large, photo-realistic indoor scene dataset of high-quality rendered images with associated ground-truth surface normals and camera poses. We use 
a pre-release subset of around 34k images. Each scene includes 3 images randomly sampled from a rendered videos.
Compared to other synthetic dataset such as SUNCG~\cite{song2016ssc}, InteriorNet has a much larger variation in camera pitch and roll. In our experiments, we randomly split InteriorNet into training, validation, and test sets using ratios of 77\%, 3\%, and 20\% based on different scene IDs. During training, we generate randomly cropped images with a vertical field of view (FoV) varying between 35 and 45 degrees. 


\medskip
\noindent
\textbf{ScanNet~\cite{dai2017scannet}} is an RGB-D video dataset containing indoor scenes with 3D camera poses and dense surface reconstructions. We use around 50K image frames, sampled approximately every second, for training and evaluation. In addition, during training we use rendered surface normals produced by Zhang and Funkhouser~\cite{zhang2018deep} for ground truth supervision, and we use the official train/val/test split based on scene id. We also use the same technique we use with InteriorNet to randomly crop images.


\medskip
\noindent\textbf{SUN360.} 
We also construct a test set of rectilinear crops from the \textbf{SUN360} panorama dataset~\cite{xiao2012recognizing} for use as a cross-dataset test set. Specifically, we extract six rectified images from each indoor panorama, with output camera parameters uniformly sampled from the ranges present in the training set of ScanNet. For all datasets and methods, we resize images to \texttt{384x288} before feeding them to the networks.



\subsection{Training details}

We implement UprightNet in PyTorch~\cite{pytorch}. For all experiments, we train using Adam~\cite{kingma2014adam}, starting from an ImageNet-pretrained encoder with initial learning rate 0.0004 and mini-batch size of 8. During training, we halve the learning rate every 5 epochs. More details on hyperparameter settings are included in the supplemental material.

\begin{figure}[t]
\centering
\begin{tabular}{c@{\hspace{0.em}}c@{\hspace{0.em}}c@{\hspace{0.em}}}
    \includegraphics[width=0.15\textwidth]{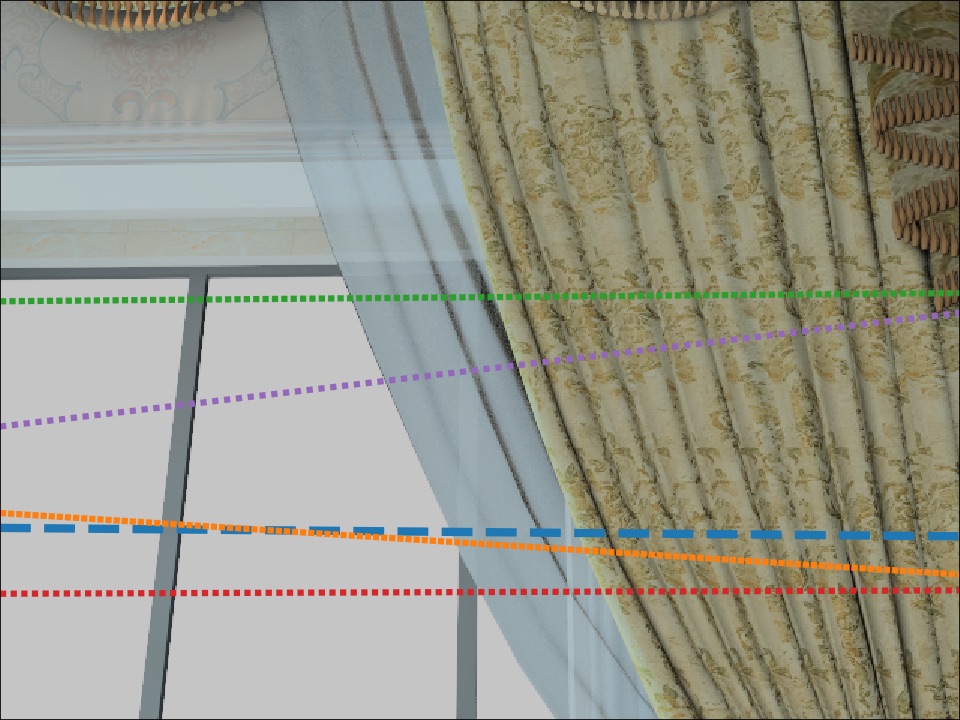} \vspace{-0.05em}& 
    \includegraphics[width=0.15\textwidth]{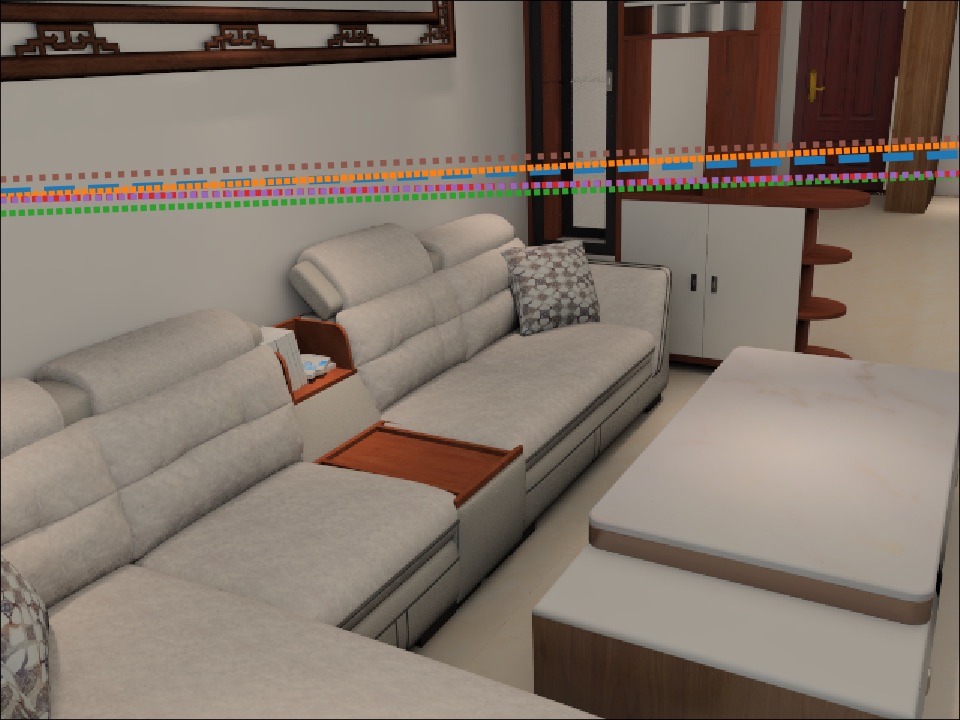} \vspace{-0.05em}&
    \includegraphics[width=0.15\textwidth]{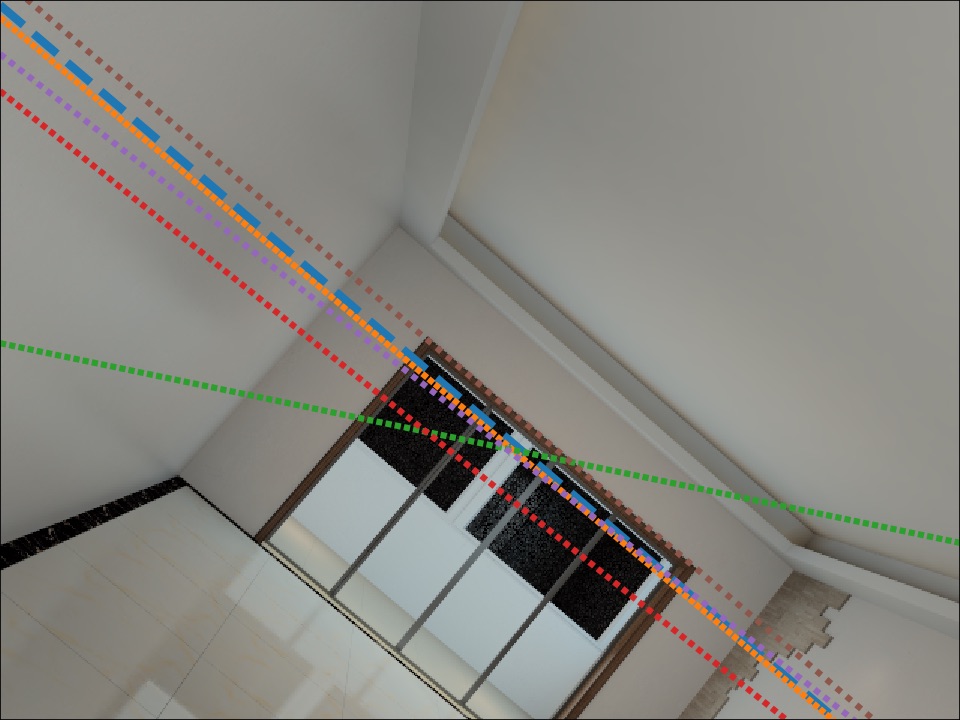} \vspace{-0.05em} \\
    \includegraphics[width=0.15\textwidth]{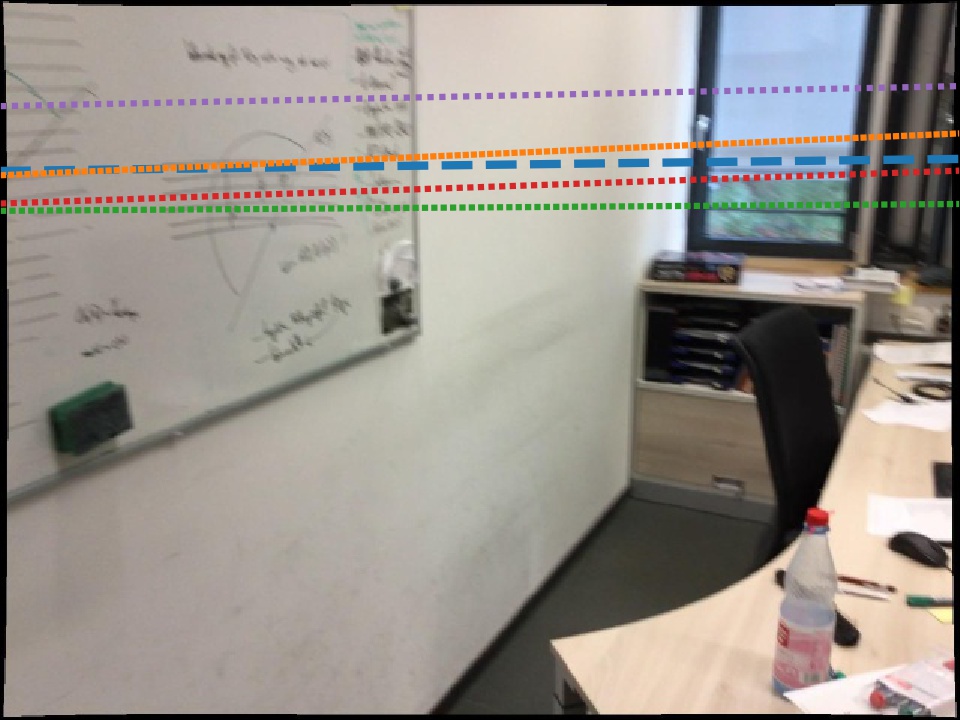} \vspace{-0.05em} & 
    \includegraphics[width=0.15\textwidth]{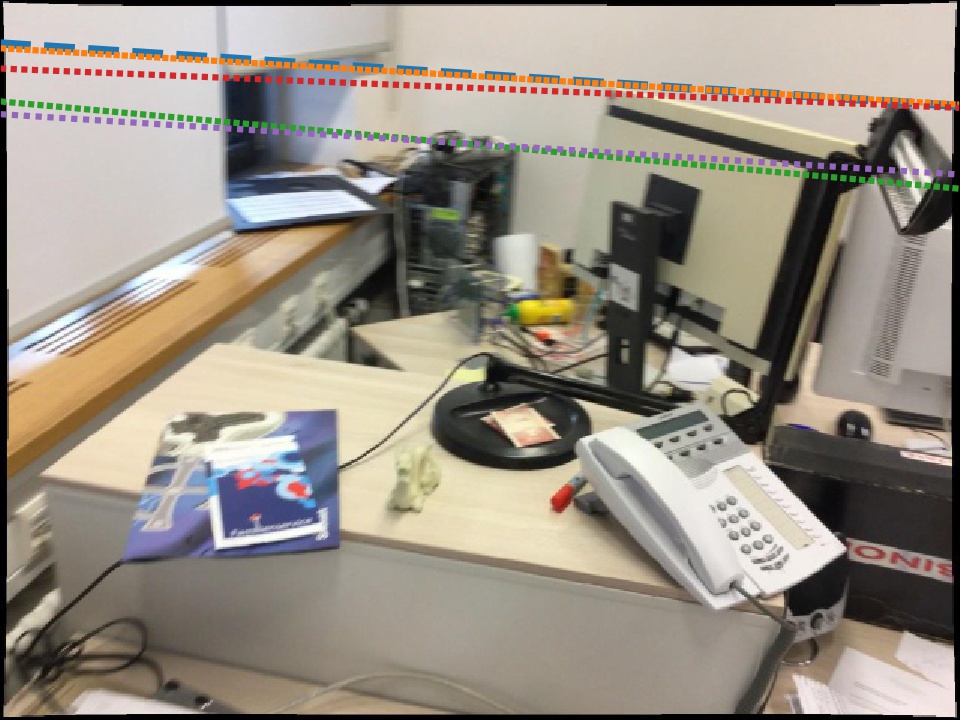} \vspace{-0.05em} &     
    \includegraphics[width=0.15\textwidth]{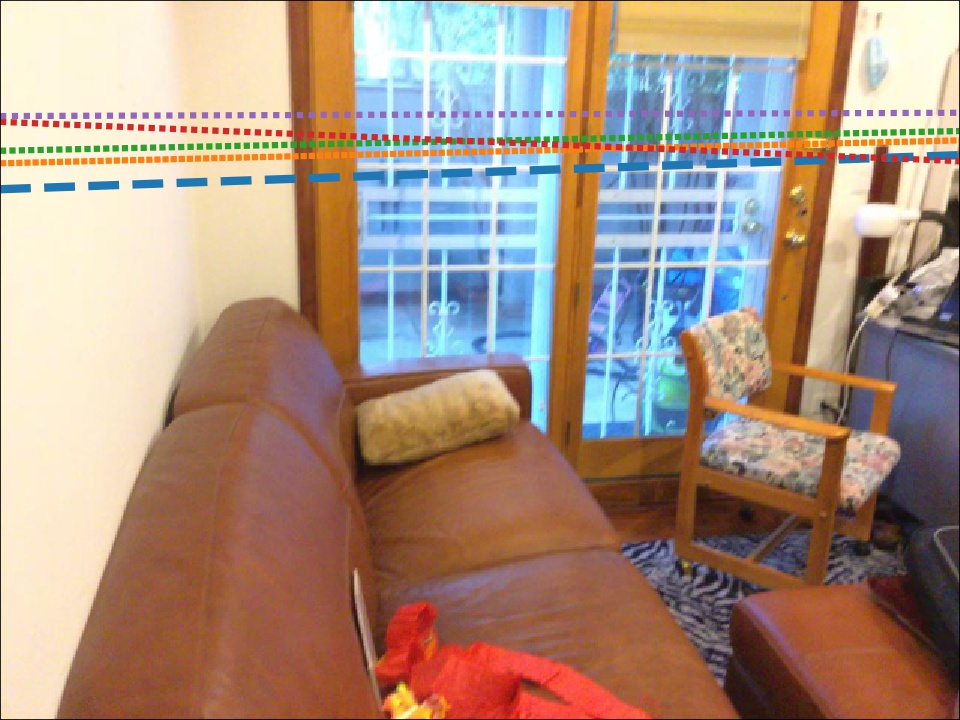} \vspace{-0.05em} \\
    \includegraphics[width=0.15\textwidth]{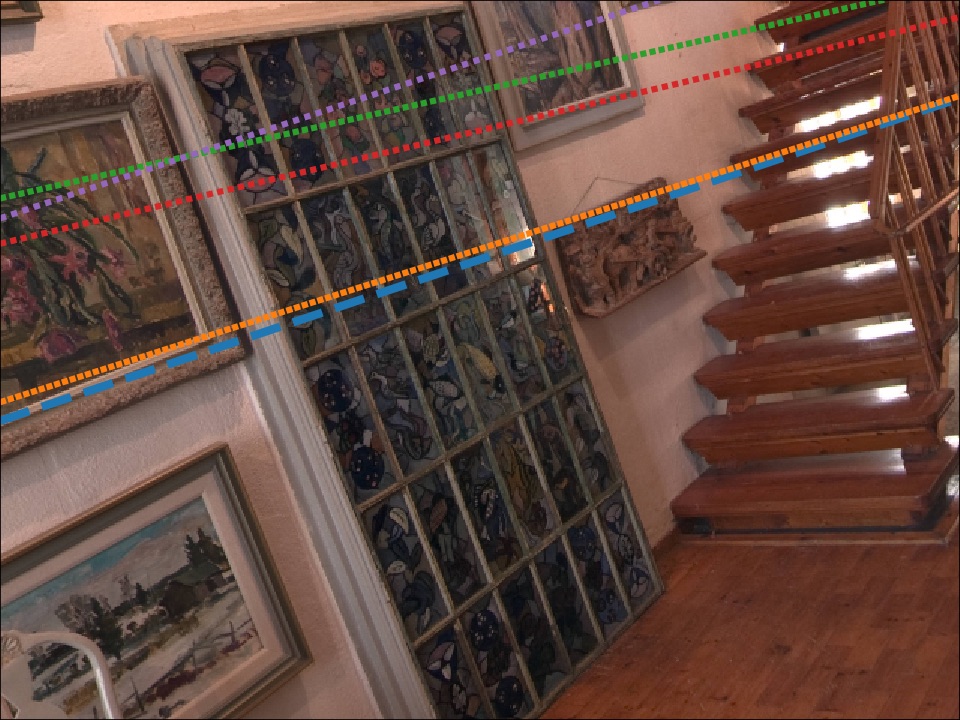} \vspace{-0.01em}& 
    \includegraphics[width=0.15\textwidth]{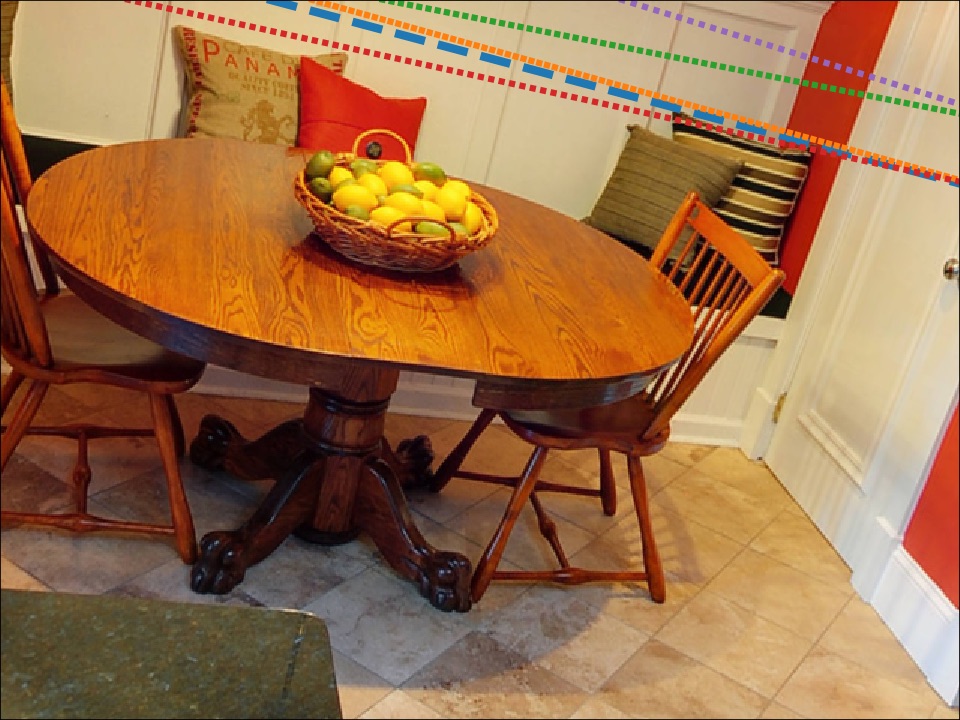} \vspace{-0.01em}&     
    \includegraphics[width=0.15\textwidth]{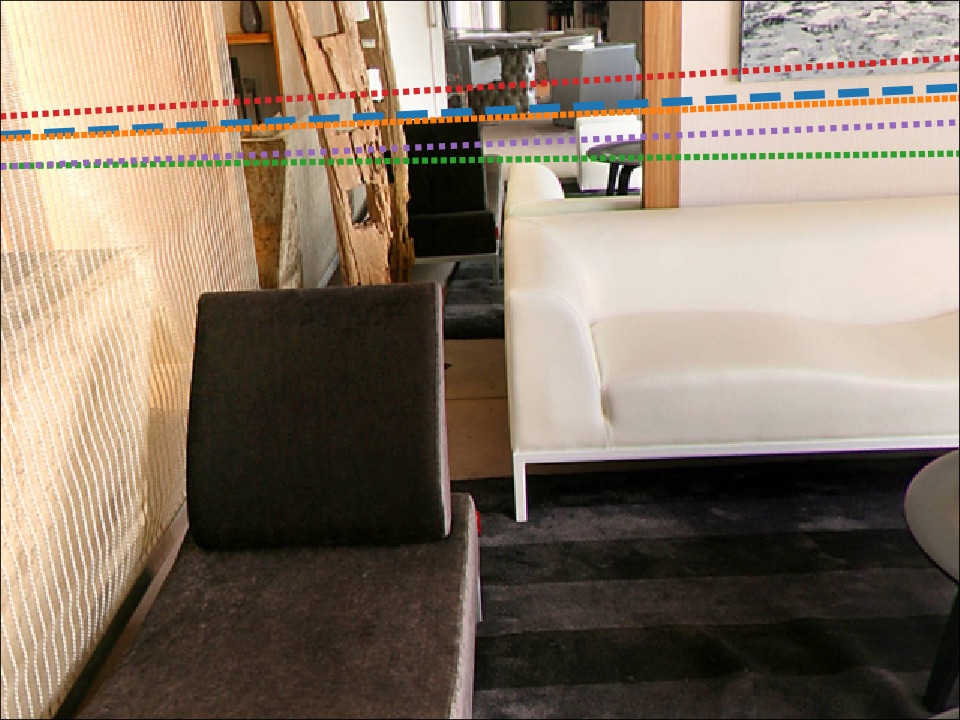} \vspace{-0.01em}
\end{tabular}
\includegraphics[width=0.45\textwidth]{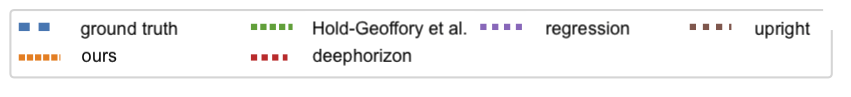}
\caption{\textbf{Qualitative comparison of horizon line predictions.} From top row to bottom row: InteriorNet, ScanNet and SUN360. Our trained model outperforms other baselines in terms of accuracy on all three datasets.}
\label{fig:examples}
\end{figure}

\subsection{Comparisons to baseline methods}

We compare UprightNet with four baseline methods:
\begin{packed_item}
  \item[$\bullet$] A regression baseline: a CNN that predicts roll and pitch angles, trained with an $\ell_1$ loss.
  \item[$\bullet$] DeepHorizon~\cite{workman2016hlw}, a CNN classification approach to predicting horizon offsets and slopes.
  \item[$\bullet$] Hold-Geoffroy~\etal~\cite{hold2018perceptual}: a CNN classification method for predicting horizon lines and fields of view.
  \item[$\bullet$] Lee~\etal~\cite{lee2014automatic}: a state-of-the-art classical geometry-based method.
\end{packed_item}
To facilitate fair comparison, we re-implemented \cite{workman2016hlw} and \cite{hold2018perceptual} based on their published descriptions. For all learning-based methods, we adopt the same network architecture and pretrained weights as our method, and train and evaluate them on the same datasets using the same training strategy.


To evaluate estimated camera orientations,
we compute the mean and median angular error between the predicted and ground truth up vectors, as in Equation~\ref{eq:4}, as well as absolute errors in both pitch and roll angles.

Table~\ref{exp:interiornet} shows quantitative results on InteriorNet, and the top row of Figure~\ref{fig:examples} shows a visualization of estimated horizon lines superimposed on the input images for qualitative comparisons. Our proposed method achieve a significant improvement (30\%) compared to prior CNN-based and geometry-based methods for all error metrics.  

Table~\ref{exp:scannet} shows quantitative results on ScanNet, and the second row of Figure~\ref{fig:examples} shows predicted horizon lines on that dataset. ScanNet is more challenging for image calibration compared to InteriorNet due to motion blur, extreme pitch angles, and more complex scene structures. Nevertheless, our method still improves calibration accuracy by a relative 20\% in angular and pitch error, and is slightly better than the best baseline methods in terms of roll angle.

\begin{figure}[t]
  \centering
    \begin{tabular}{@{\hspace{-0.1em}}c@{\hspace{-0.1em}}c@{\hspace{-0.1em}}c@{\hspace{-0.1em}}c@{\hspace{-0.1em}}}
        \includegraphics[width=0.24\columnwidth]{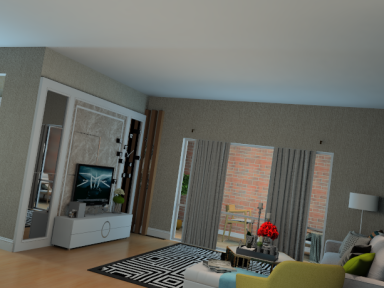} \vspace{-0.02em} & 
        \includegraphics[width=0.24\columnwidth]{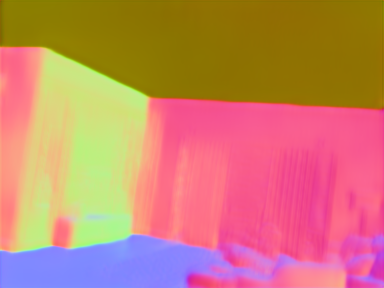}  \vspace{-0.02em} &
       \includegraphics[width=0.24\columnwidth]{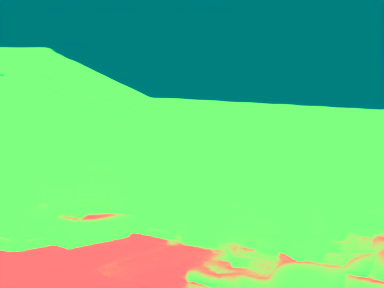}  \vspace{-0.02em} &
        \includegraphics[width=0.24\columnwidth]{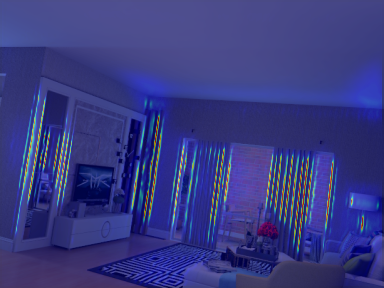}  
        \vspace{-0.02em} \\ 
        \includegraphics[width=0.24\columnwidth]{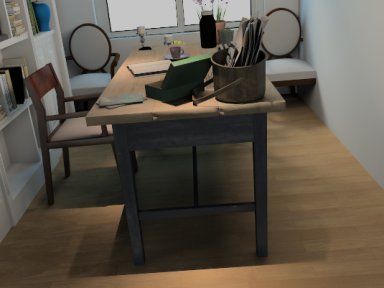} \vspace{-0.02em} & 
        \includegraphics[width=0.24\columnwidth]{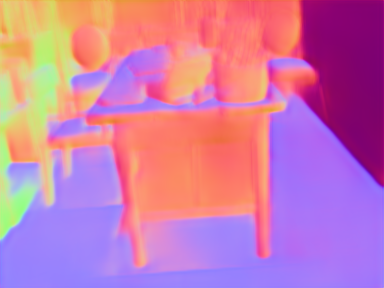}  \vspace{-0.02em} &
       \includegraphics[width=0.24\columnwidth]{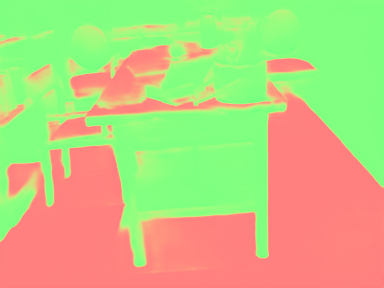}  \vspace{-0.02em} &
        \includegraphics[width=0.24\columnwidth]{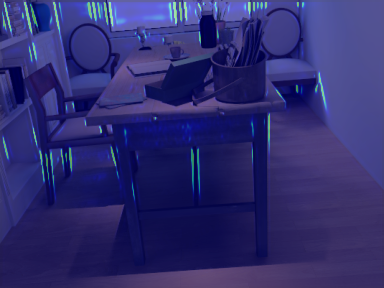} 
        \vspace{-0.02em} \\     
        \includegraphics[width=0.24\columnwidth]{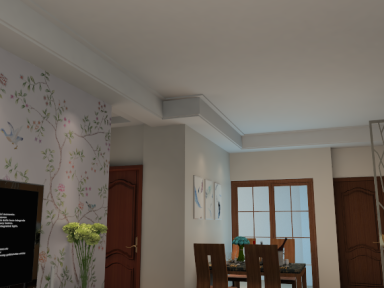} \vspace{-0.02em} & 
        \includegraphics[width=0.24\columnwidth]{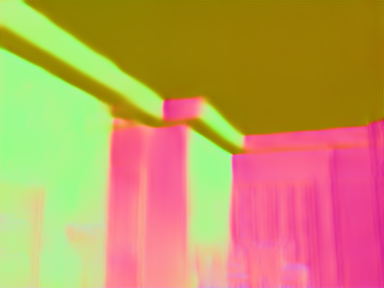}  \vspace{-0.02em} &
 
      \includegraphics[width=0.24\columnwidth]{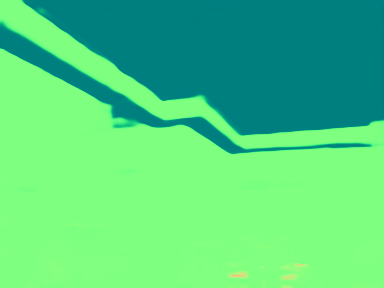}  \vspace{-0.02em} &
        \includegraphics[width=0.24\columnwidth]{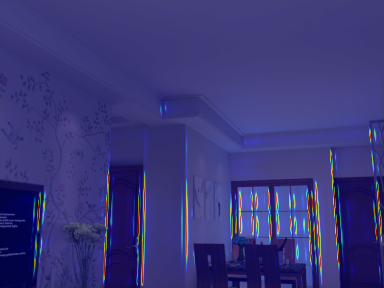} 
        \vspace{-0.02em} \\     
        \includegraphics[width=0.24\columnwidth]{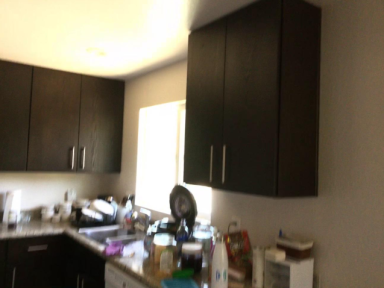} \vspace{-0.02em} & 
        \includegraphics[width=0.24\columnwidth]{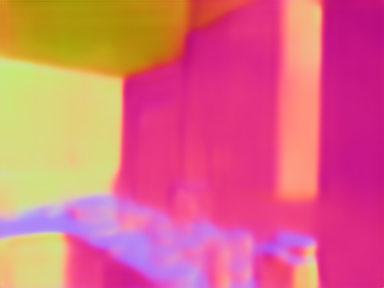}  \vspace{-0.02em} &
       \includegraphics[width=0.24\columnwidth]{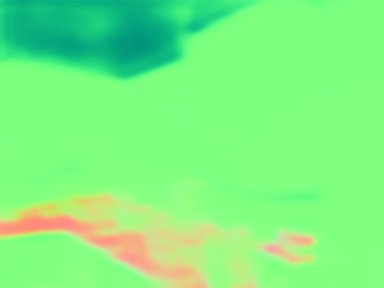}  \vspace{-0.02em} &
        \includegraphics[width=0.24\columnwidth]{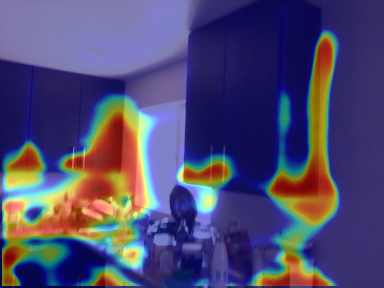}  \vspace{-0.02em} 
         \\    
        \includegraphics[width=0.24\columnwidth]{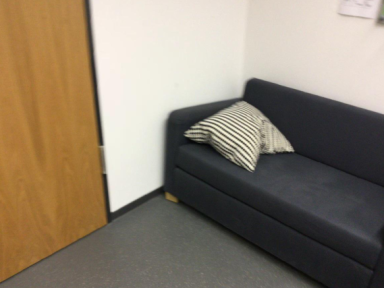} \vspace{-0.02em} & 
        \includegraphics[width=0.24\columnwidth]{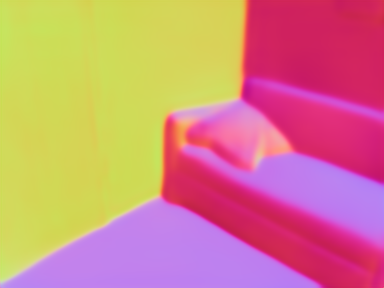}  \vspace{-0.02em} &
       \includegraphics[width=0.24\columnwidth]{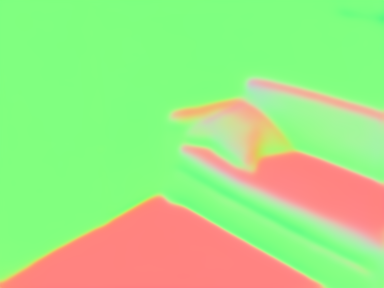}  \vspace{-0.02em} &
        \includegraphics[width=0.24\columnwidth]{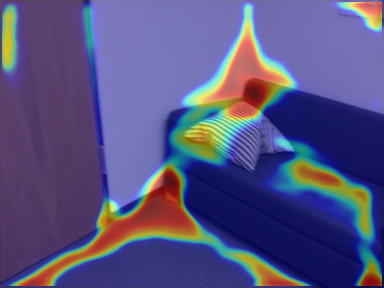}  \vspace{-0.02em} 
         \\    
        \includegraphics[width=0.24\columnwidth]{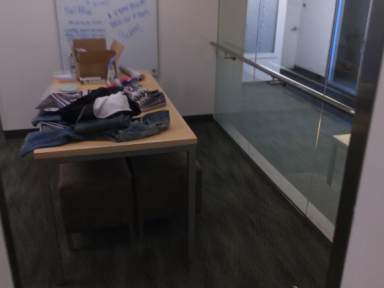} \vspace{-0.02em} &
        \includegraphics[width=0.24\columnwidth]{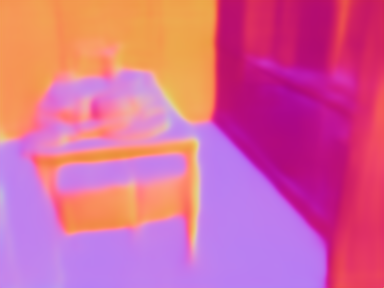}  \vspace{-0.02em} &
       \includegraphics[width=0.24\columnwidth]{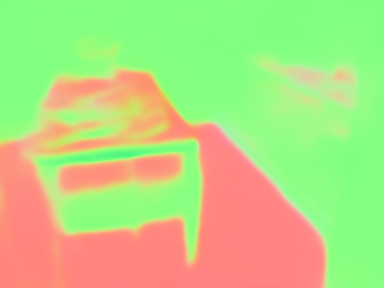}  \vspace{-0.02em} &
        \includegraphics[width=0.24\columnwidth]{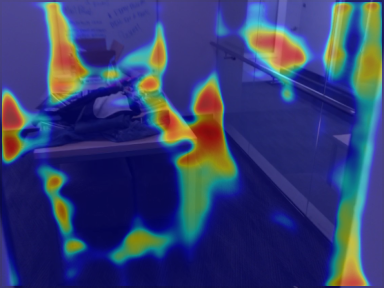}  \vspace{-0.02em} 
         \\    
        (a) Image \vspace{-0.1em} & 
        (b) $\normal^c$ \vspace{-0.1em} & 
        (c) $\upFz$ \vspace{-0.1em} &
        (d) weights \vspace{-0.1em} 
    \end{tabular} 
  \caption{\textbf{Visualizations of predictions} in InteriorNet (top 3 rows) and ScanNet (bottom 3 rows). In (d), we overlay weight maps (combined weights for $\normal$, $\tangent$ and $\bitangent$) over input images. Blue=small weights, red=large weights.}
  \label{fig:interiornet_pred} \vspace{-0.5em}
\end{figure}

Figure~\ref{fig:interiornet_pred} shows  visualizations of our predicted $\normal^c$, $\upFz$, and weights for InteriorNet and ScanNet test images. For visualization, in Figure~\ref{fig:interiornet_pred}(d) we combine the three weights maps by adding the normalizing them, and overlay them on the original input images. Interestingly, the network learns weight maps that tend to focus on the vicinity of vertical/horizontal surface line junctions. Such lines are very strong cues for vanishing points. For instance, $\tangent^c$ in the vicinity of vertical junctions and $\bitangent^c$ in the vicinity of horizontal junctions both represent 3D directions of their respective junction lines, and also point towards 2D vanishing points~\cite{hartley2003multiple}. 
It is interesting that the network ``discovers'' such features automatically

\begin{table}[tb]
\resizebox{\columnwidth}{!}{%
\begin{tabular}{lccccccccc} 
\toprule
Method
& \multicolumn{2}{c}{angular error $(^{\circ})$} & \multicolumn{2}{c}{pitch error $(^{\circ})$} & \multicolumn{2}{c}{roll error $(^{\circ})$}\\
\cmidrule(lr){2-3} \cmidrule(lr){4-5} \cmidrule(lr){6-7}
& avg. & med. & avg. & med. & avg. & med. \\ \midrule

CNN Regression  &  2.29  &   1.20      &   1.88   &   1.02 & 0.93 & 0.34   \\
Upright~\cite{lee2014automatic}  & 3.95 & 1.76 & 3.59	& 1.70 & 1.15	& 0.33		 \\
DeepHorizon~\cite{workman2016hlw} &  2.12 &	1.13 & 1.65	& 0.85	& 0.93	& 0.40	 \\
Hold-Geoffroy~\etal~\cite{hold2018perceptual} & 1.76	& 0.81& 1.35	&0.58	&0.81 &0.26    \\ 
 \midrule
Ours  &   \textbf{1.17} & \textbf{0.52} & \textbf{0.99} & \textbf{0.47}  & \textbf{0.44} &	\textbf{0.11} 	    \\
\bottomrule
\end{tabular}
}
\caption{\textbf{Quantitative comparisons on the InteriorNet test set.} Our approach has the best performance in terms of all the metrics. All errors are in degrees and lower is better.}
\label{exp:interiornet}
\end{table}

\begin{table}[tb]
\resizebox{\columnwidth}{!}{%
\begin{tabular}{lccccccccc} 
\toprule
Method
& \multicolumn{2}{c}{angular error $(^{\circ})$} & \multicolumn{2}{c}{pitch error $(^{\circ})$} & \multicolumn{2}{c}{roll error $(^{\circ})$}\\
\cmidrule(lr){2-3} \cmidrule(lr){4-5} \cmidrule(lr){6-7}
& avg. & med. & avg. & med. & avg. & med. \\ \midrule

CNN Regression  &  4.88 &	3.63	 & 3.85 &	2.59 &	2.24	& 1.57 \\
Upright~\cite{lee2014automatic} & 14.51 &	5.71	& 4.65	& 2.13	& 12.99	& 4.32 \\
DeepHorizon~\cite{workman2016hlw} & 5.05 & 3.80 	& 3.81	& 2.56	& 2.51	& 1.82 \\
Hold-Geoffroy~\etal~\cite{hold2018perceptual} & 4.55 &	3.33  & 3.53	& 2.33		& 2.15	& 1.50	   \\ 
 \midrule
Ours  &   \textbf{3.74} & \textbf{2.94}  & \textbf{2.88} & \textbf{2.04}  & \textbf{2.12} &	\textbf{1.48} 	    \\
\bottomrule
\end{tabular}
}
\caption{\textbf{Quantitative comparisons on the ScanNet testset.} See Table~\ref{exp:interiornet} for descriptions.}
\label{exp:scannet}
\end{table}

\subsection{Ablation analysis}

We now explore how the weight maps and different configurations of surface geometry supervision affect the performance of UprightNet through validation on InteriorNet.  

\medskip
\noindent
\textbf{Impact of weights.}
We wanted to see if the network learns to correctly assign large weights to regions with smaller alignment error. We compute an alignment score map by using the ground-truth up vector $\upvector$ to align each column of $\hat{\mathbf{F}}^c$ with its corresponding scalar in $\hat{\mathbf{f}}_z^c$. We define  alignment scores for surface normals as 
$S_{\normal} = \exp (- 10 |\upvector^T  \hat{\normal}^c - \hat{n}^g_z|)$ and similarly for the two tangent vectors.

\begin{figure}[t]
  \centering
    \begin{tabular}{@{\hspace{-0.1em}}c@{\hspace{-0.1em}}c@{\hspace{-0.1em}}c@{\hspace{-0.1em}}c@{\hspace{-0.1em}}}
        \includegraphics[width=0.24\columnwidth]{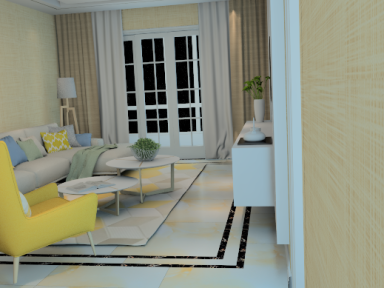} \vspace{-0.05em} & 
        \includegraphics[width=0.24\columnwidth]{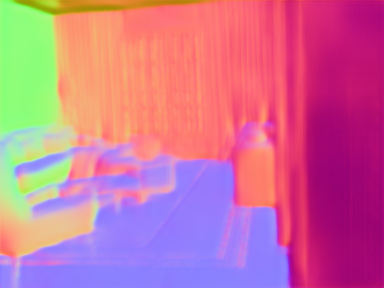}  \vspace{-0.05em} &
        \includegraphics[width=0.24\columnwidth]{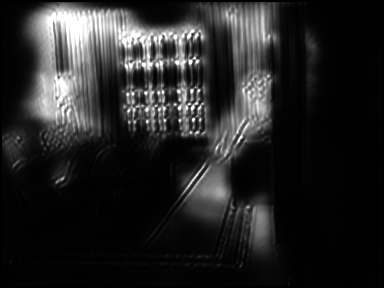}  \vspace{-0.05em} &
        \includegraphics[width=0.24\columnwidth]{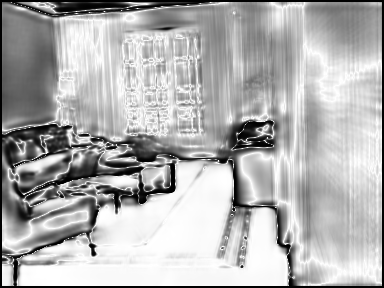}  \vspace{-0.05em} \\
        {\small (a) Image} & {\small (b) $\normal^c$} & {\small
          (c) $w_{\normal}$} & {\small (d) $S_{\normal}$} \\ 
         \includegraphics[width=0.24\columnwidth]{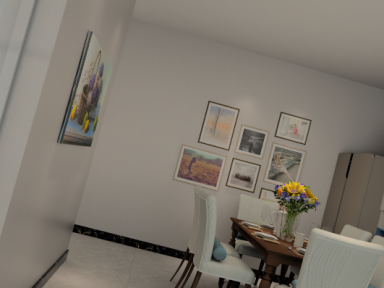} \vspace{-0.05em} & 
        \includegraphics[width=0.24\columnwidth]{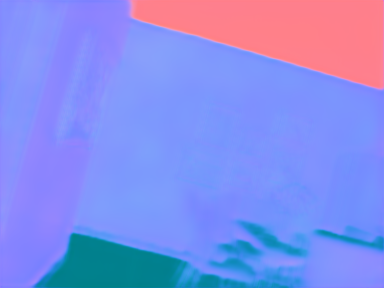}  \vspace{-0.05em} &
        \includegraphics[width=0.24\columnwidth]{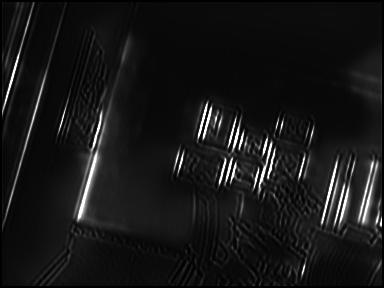}  \vspace{-0.05em} &
        \includegraphics[width=0.24\columnwidth]{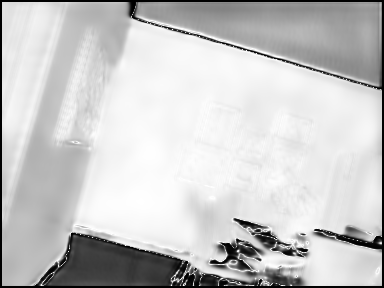} \vspace{-0.05em} \\ 
        {\small (a) Image} & {\small (b) $\tangent^c$} & {\small
          (c) $w_{\tangent}$} & {\small
          (d) $S_{\tangent}$} \\
        \includegraphics[width=0.24\columnwidth]{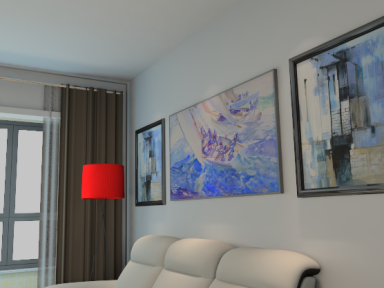} \vspace{-0.05em} & 
        \includegraphics[width=0.24\columnwidth]{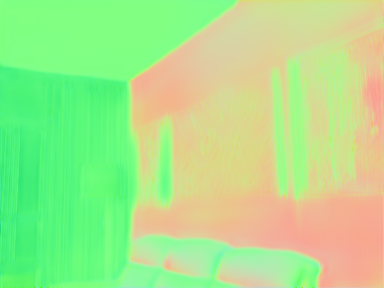}  \vspace{-0.05em} &
        \includegraphics[width=0.24\columnwidth]{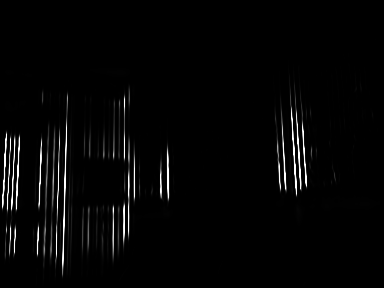}  \vspace{-0.05em} &
        \includegraphics[width=0.24\columnwidth]{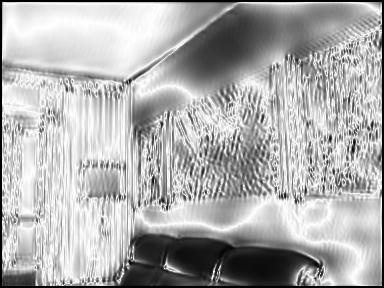} \vspace{-0.05em} \\ 
        {\small (a) Image} & {\small (b) $\bitangent^c$} & {\small
          (c) $w_{\bitangent}$} & {\small
          (d) $S_{\bitangent}$}
    \end{tabular} 
  \caption{\textbf{Comparisons between predicted weights and alignment scores.}  From left to right (a) input image, (b) predict camera surface frames, (c) predicted weights, (d) alignment scores $S$. In (c) and (d), white indicates large weights and high alignment score (i.e.\ low alignment error).}\label{fig:alignment}
\end{figure}

Figure~\ref{fig:alignment} visualizes these alignment scores, and compares them with the predicted weights. The network indeed tends to predict large weights where the alignment error is small. This suggests that, while our surface geometry predictions are not always accurate, the weights can capture which image regions are more likely to have correct predictions, thus leading to better estimation of camera orientation. 


\medskip
\noindent
\textbf{Impact of camera orientation loss.}
As shown in Table~\ref{exp:ablation_1}, we evaluate our models using different configurations to analyze their influence on the performance. Comparing with direct estimation of camera pose from the predicted surface frame correspondences, end-to-end optimization with camera orientation loss boosts performance significantly. We observe an additional increase in performance by incorporating predicted weights into orientation estimation. Since the supervision of surface frames in both coordinate systems is not required during training, we can also train a model using supervision only from local camera frames, or only from global upright frames. The ablations shown in Table~\ref{exp:ablation_1} suggest that using both local camera and global upright surface geometry as supervision leads to the best performance.

\medskip
\noindent
\textbf{Impact of surface geometry representation.}
We also explore the influence of different surface geometry representations on camera orientation estimation. In particular, we compare our full surface geometry representation to (1) a single vector representation, i.e., just one of $ \left(\normal, \tangent, \bitangent \right)$, and (2) a combination of any two of them. As shown in Table~\ref{exp:ablation_2}, our proposed full surface frame representation achieves the best performance. This suggests that each basis vector of our surface frames captures complementary geometry cues in indoor scenes. In particular, $\normal^c$ of grounds/ceilings and $\tangent^c$ of vertical lines directly represent the scene up vectors we seek, while $\bitangent^c$ of horizontal lines on supporting surfaces could correspond to major vanishing points in the scene.


\begin{table}[tb]
\resizebox{\columnwidth}{!}{%
\begin{tabular}{lccccccccc} 
\toprule
Method
& \multicolumn{2}{c}{angular error $(^{\circ})$} & \multicolumn{2}{c}{pitch error $(^{\circ})$} & \multicolumn{2}{c}{roll error $(^{\circ})$}\\
\cmidrule(lr){2-3} \cmidrule(lr){4-5} \cmidrule(lr){6-7}
& avg. & med. & avg. & med. & avg. & med. \\ \midrule

Ours (w/o weight) & 1.83 &	1.15 & 1.22 &	0.81  &	0.90&	0.57 \\
Ours (w/o $\Lpose$) & 2.71 &	1.74  &	1.21 & 2.83  	& 1.37	& 0.93		\\
Ours (w/o $F^c$ loss) & 1.70 & 1.11 & 1.31 & 0.99  & 0.76 & 0.57 \\
Ours (w/o $F^u$ loss) & 1.82 & 1.18 & 1.32 & 0.98 & 0.79 & 0.55   \\
\midrule
Ours (full)  &   \textbf{1.17} & \textbf{0.52}  & \textbf{0.99} & \textbf{0.47} & \textbf{0.44} &	\textbf{0.11} 	 \\
\bottomrule
\end{tabular}
}
\caption{\textbf{Ablation studies for different configurations on the InteriorNet test set.} Ours (full) indicate our proposed full method, which achieves the best performance among all tested configurations.}
\label{exp:ablation_1}
\end{table}

\begin{table}[tb]
\resizebox{\columnwidth}{!}{%
\begin{tabular}{lccccccccc} 
\toprule
Method
& \multicolumn{2}{c}{angular error $(^{\circ})$} & \multicolumn{2}{c}{pitch error $(^{\circ})$} & \multicolumn{2}{c}{roll error $(^{\circ})$}\\
\cmidrule(lr){2-3} \cmidrule(lr){4-5} \cmidrule(lr){6-7}
& avg. & med. & avg. & med. & avg. & med. \\ \midrule
$\normal$ &  1.81	& 1.13	& 1.35 &	0.85  & 0.90 & 0.45  \\
$\tangent$ &  2.11	&1.28 & 1.60	& 0.96 &	1.06	& 0.47	   \\
$\bitangent$ &  1.82 & 1.09  & 1.39	& 0.84	& 0.81	& 0.47 	   \\
\midrule
$\normal\tangent$  & 1.52 & 0.97 & 1.27 & 0.83  &	0.58 & 0.29  \\
$\normal\bitangent$ & 1.17	& 0.89 & 1.33 & 	0.77  & 0.78&	0.29		 	 \\
$\tangent\bitangent$ & 1.56	& 0.66	& 1.29	& 0.62 & 0.66	& 0.20	  \\
 \midrule
$\normal\tangent\bitangent$  &   \textbf{1.17} & \textbf{0.52} & \textbf{0.99} & \textbf{0.47}  & \textbf{0.44} &	\textbf{0.11}	 \\
\bottomrule
\end{tabular}
}
\caption{\textbf{Ablation studies for different surface geometry representations on the InteriorNet test set.} $\normal\tangent\bitangent$ indicates our full surface geometry representation. }
\label{exp:ablation_2}
\vspace{-0.5em}
\end{table}

\subsection{Generalization to SUN360} \label{sec:sun360}
\begin{figure}[t]
  \centering
    \begin{tabular}{@{\hspace{-0.1em}}c@{\hspace{-0.1em}}c@{\hspace{-0.1em}}c@{\hspace{-0.1em}}c@{\hspace{-0.1em}}}
        \includegraphics[width=0.24\columnwidth]{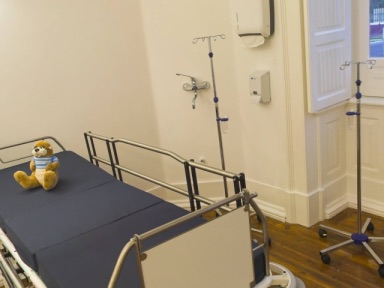} \vspace{-0.02em} & 
        \includegraphics[width=0.24\columnwidth]{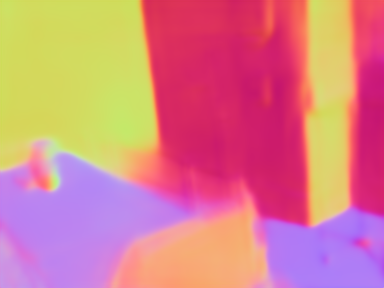}  \vspace{-0.02em} &
       \includegraphics[width=0.24\columnwidth]{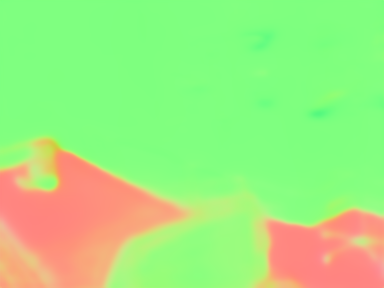}  \vspace{-0.02em} &
        \includegraphics[width=0.24\columnwidth]{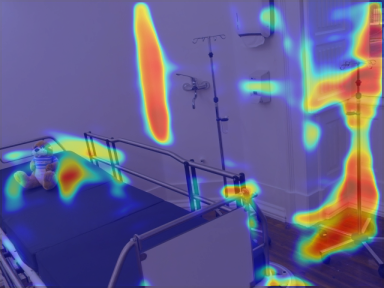}  
        \vspace{-0.02em} \\ 
        \includegraphics[width=0.24\columnwidth]{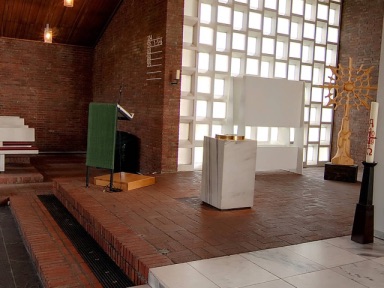} \vspace{-0.02em} & 
        \includegraphics[width=0.24\columnwidth]{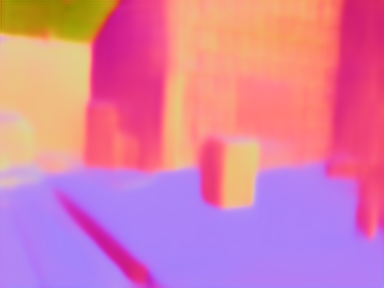}  \vspace{-0.02em} &
       \includegraphics[width=0.24\columnwidth]{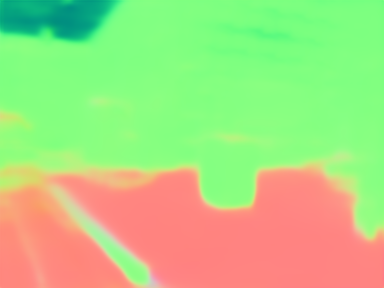}  \vspace{-0.02em} &
        \includegraphics[width=0.24\columnwidth]{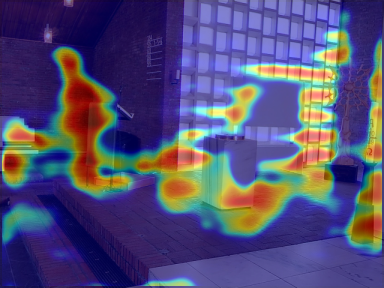} 
        \vspace{-0.02em} \\     
        \includegraphics[width=0.24\columnwidth]{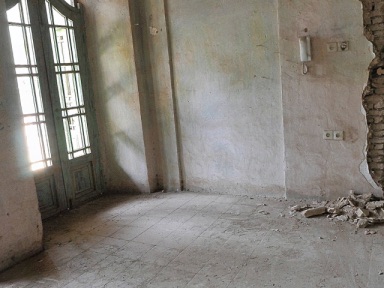} \vspace{-0.02em} & 
        \includegraphics[width=0.24\columnwidth]{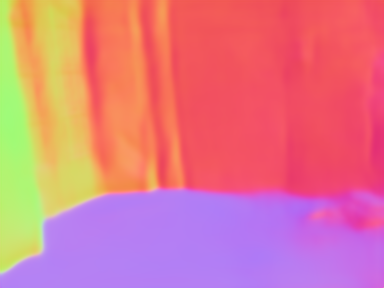}  \vspace{-0.02em} &
       \includegraphics[width=0.24\columnwidth]{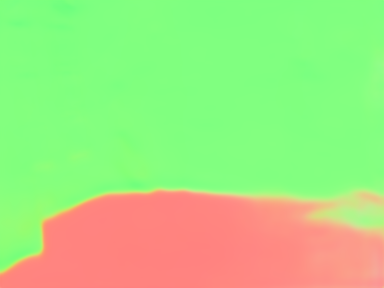}  \vspace{-0.02em} &
        \includegraphics[width=0.24\columnwidth]{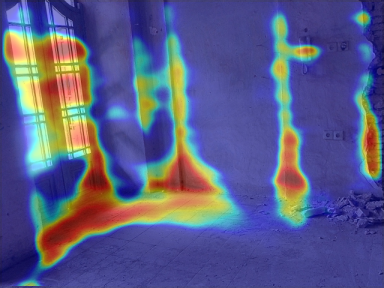} 
        \vspace{-0.02em} \\   
        \includegraphics[width=0.24\columnwidth]{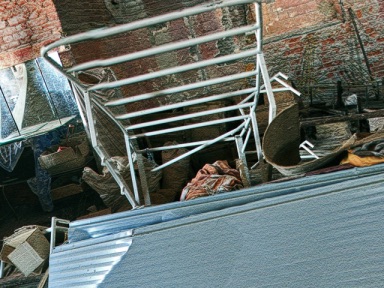} \vspace{-0.02em} & 
        \includegraphics[width=0.24\columnwidth]{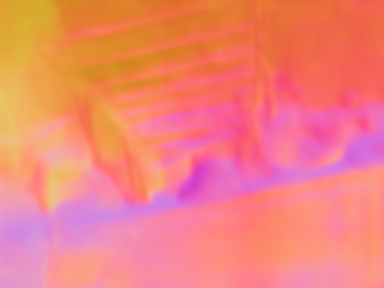}  \vspace{-0.02em} &
       \includegraphics[width=0.24\columnwidth]{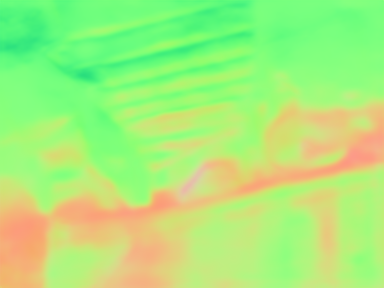}  \vspace{-0.02em} &
        \includegraphics[width=0.24\columnwidth]{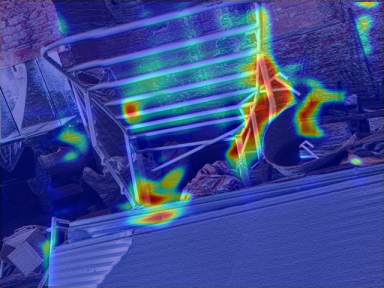} 
        \vspace{-0.02em} \\   
        (a) Image \vspace{-0.05em} & 
        (b) $\normal^c$ \vspace{-0.05em} & 
        (c) $\upFz$ \vspace{-0.05em} & 
        (d) weights \vspace{-0.05em} 
    \end{tabular} 
  \caption{\textbf{Visualizations of predictions on the SUN360 testset.} The last row shows a failure case.} \vspace{-0.5em}
  \label{fig:sun36}
\end{figure}

We explore the ability of different learning-based calibration methods to generalize across datasets by taking models trained on ScanNet and testing them on crops from the SUN360 indoor panorama dataset. We summarize the results in Table~\ref{exp:cross}, and visualize surface normal and weight predictions of our model in Figure~\ref{fig:sun36}. While the errors are naturally higher on this unseen dataset, UprightNet still outperforms prior methods by large margins in all error metrics. This suggests that, compared with implicit features learned from direct regression or classification, using an intermediate geometric representation can help a model attain better generalization to new distributions of indoor imagery.

\begin{figure}[t]
  \centering
    \begin{tabular}{@{\hspace{0em}}c@{\hspace{0em}}c@{\hspace{0em}}c@{\hspace{0em}}}
        \includegraphics[width=0.3\columnwidth]{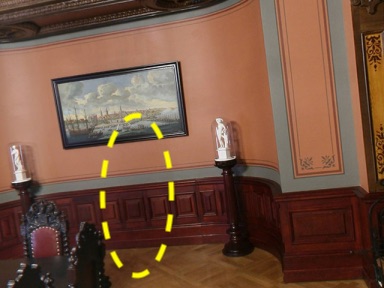} \vspace{-0.05em} & 
        \includegraphics[width=0.3\columnwidth]{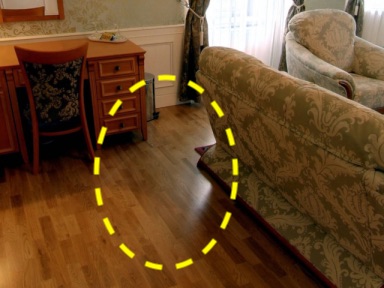}  \vspace{-0.05em} &
        \includegraphics[width=0.3\columnwidth]{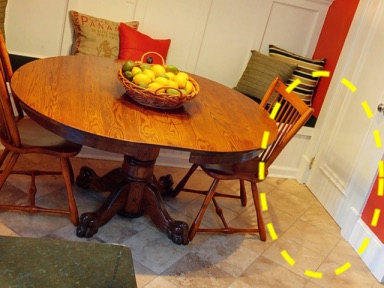} \vspace{-0.05em}  \\ 
        \includegraphics[width=0.3\columnwidth]{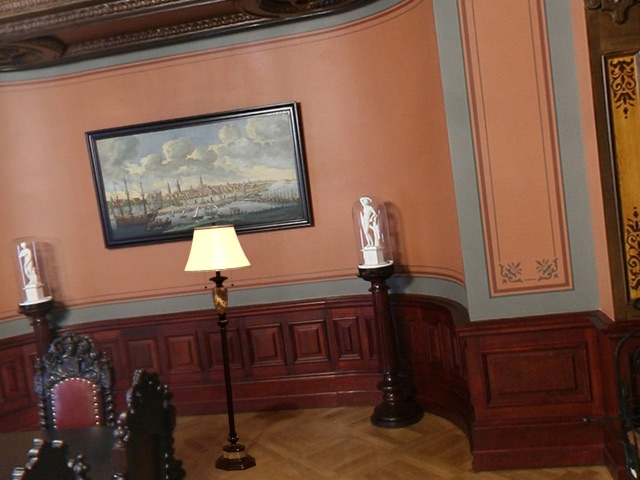} \vspace{-0.05em} & 
        \includegraphics[width=0.3\columnwidth]{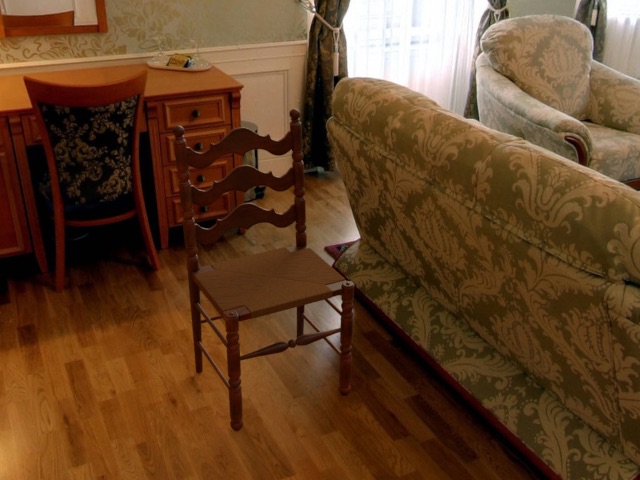}  \vspace{-0.05em} &
        \includegraphics[width=0.3\columnwidth]{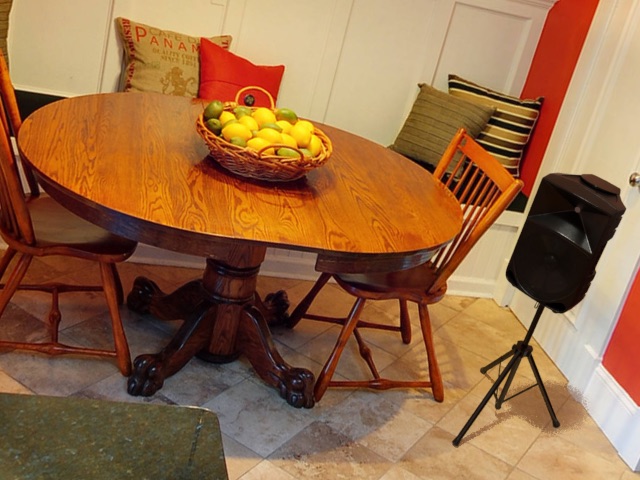} \vspace{-0.05em}  \\ 
    \end{tabular} \vspace{-0.5em}
  \caption{\textbf{Application.} Virtual insertion of 3D objects in images from SUN360 dataset using the camera orientation estimated by UprightNet.}
  \label{fig:demo}
\end{figure}

We also demonstrate the use of UprightNet for an application in virtual object insertion in Figure~\ref{fig:demo}. We orient a 3D object with the camera's pitch and roll estimated by UprightNet. The 2D translation of the object along the ground plane is manually chosen by the artist. Finally, a 2D render of the object is composited on top of the image to produce a final results, shown in the bottom row of Figure~\ref{fig:demo}.  

\begin{table}[tb]
\resizebox{\columnwidth}{!}{%
\begin{tabular}{lccccccccc} 
\toprule
Method
& \multicolumn{2}{c}{angular error $(^{\circ})$} & \multicolumn{2}{c}{pitch error $(^{\circ})$} & \multicolumn{2}{c}{roll error $(^{\circ})$}\\
\cmidrule(lr){2-3} \cmidrule(lr){4-5} \cmidrule(lr){6-7}
& avg. & med. & avg. & med. & avg. & med. \\ \midrule
CNN Regression  &  10.43	& 6.99 & 9.57 &	6.10	 & 3.10	& 2.17		 \\
DeepHorizon~\cite{workman2016hlw} & 9.53 &	6.28	& 8.68	&5.50& 2.98	& 1.89		 \\
Hold-Geoffroy~\etal~\cite{hold2018perceptual} & 10.41 &	6.92	 &9.57	&6.09 	& 3.11	 &2.20	    \\ 
 \midrule
Ours  &   \textbf{7.81} & \textbf{5.53} & \textbf{7.59}  & \textbf{4.94}  & \textbf{2.30} &	\textbf{1.53}	    \\
\bottomrule
\end{tabular}
}
\caption{\textbf{Generalization performance on the SUN360 test set.} All methods are trained/validated on ScanNet. Our method achieves the best performance on all metrics.}
\label{exp:cross}
\end{table}

\section{Limitations and future work}
 The last row of Figure~\ref{fig:sun36} demonstrate a common failure mode of our approach, where the image lacks sufficient supporting structure for the network to reason about geometry, resulting in inaccurate surface geometry and camera orientation predictions. In future work, other explicit 2D geometric priors, such as vanishing points, or camera intrinsics (if known) can also be integrated to our framework. 
\section{Conclusion}

We introduced UprightNet, a new method for predicting 2DoF camera orientation from a single indoor image. Our key insight is to leverage surface geometry information from both the camera and upright coordinate systems, and pose camera orientation prediction as an alignment problem between these two frames.  In particular, we showed how a network can be trained to predict camera-centric and global surface frames, and combine them with weights to estimate the camera orientation. Our evaluations demonstrated not only more accurate orientation estimation, but also better generalization to unseen datasets, compared with prior state-of-the-art methods.


\medskip
\noindent \textbf{Acknowledgments.}
We thank Geoffrey Oxholm, Qianqian Wang, and Kai Zhang for helpful discussion and comments. This work was funded in part by the National Science Foundation (grant IIS-1149393), and by the generosity of Eric and Wendy Schmidt by recommendation of the Schmidt Futures program.




{\small
\bibliographystyle{ieee_fullname}
\bibliography{refs}
}

\end{document}